\documentclass[letterpaper, 10 pt, conference]{ieeeconf}  

\IEEEoverridecommandlockouts                              
 
\usepackage{url}
\usepackage{cite}
\usepackage{times}
\usepackage{graphicx}
\usepackage{amsmath}
\usepackage{amssymb}
\usepackage{enumerate}
\usepackage{mathtools}
\usepackage{amsmath}
\usepackage{algorithm}  
\usepackage{algorithmic}

\usepackage{subfigure}
\usepackage{float}
\usepackage{multirow}

\title{\LARGE \bf
Flow-Motion and Depth Network for Monocular Stereo and Beyond}
 
\author{Kaixuan Wang and Shaojie Shen
\thanks{All authors are with  ECE, HKUST, Hong Kong, China. {\tt\small \{kwangap, eeshaojie\}@ust.hk}.
}}
 
\begin{document}

\maketitle
\thispagestyle{empty} 
\pagestyle{empty}
 
 

\begin{abstract}
We propose a learning-based method\footnote{https://github.com/HKUST-Aerial-Robotics/Flow-Motion-Depth} that solves monocular stereo and can be extended to fuse depth information from multiple target frames. Given two unconstrained images from a monocular camera with known intrinsic calibration, our network estimates relative camera poses and the depth map of the source image. The core contribution of the proposed method is threefold. First, a network is tailored for static scenes that jointly estimates the optical flow and camera motion. By the joint estimation, the optical flow search space is gradually reduced resulting in an efficient and accurate flow estimation. Second, a novel triangulation layer is proposed to encode the estimated optical flow and camera motion while avoiding common numerical issues caused by epipolar. Third, beyond two-view depth estimation, we further extend the above networks to fuse depth information from multiple target images and estimate the depth map of the source image. To further benefit the research community, we introduce tools to generate photorealistic structure-from-motion datasets such that deep networks can be well trained and evaluated. The proposed method is compared with previous methods and achieves state-of-the-art results within less time. Images from real-world applications and Google Earth are used to demonstrate the generalization ability of the method.
\end{abstract}

\section{INTRODUCTION}

Due to the rich information in images, structure-from-motion (SfM) is of vital importance in computer vision and robotics. Given a set of unconstrained images, SfM aims to estimate the depth maps and the relative camera poses. Traditional systems, for example, COLMAP~\cite{sfm_revist, schonberger_2016}, first estimate the relative poses of cameras by finding correspondences of sparse feature points and then use the estimated camera pose to calculate dense depth maps. The extracted sparse features ignore other information in the images, such as lines, and does not contribute to the following dense depth estimation. Scene priors such as structures and object shapes are also hard to be integrated into the pipeline of traditional methods.
 
To better utilize image information and exploit context priors, many methods~\cite{demon, ls_net, BA-Net} have been proposed to solve monocular stereo (two-view SfM) problems using convolutional neural networks (CNNs). DeMoN~\cite{demon} is a pioneering work that first estimates an optical flow and then decomposes it into a depth map and camera pose. The optical flow, depth maps, and camera poses are then iteratively refined by a chain of encoder-decoder networks to handle large viewing angles. LS-Net~\cite{ls_net} uses a predicted depth map and camera pose as the initialization to iteratively minimize the photometric reprojection error through a learning-based solver. Different from LS-Net where the update steps are computed by a network, BA-Net~\cite{BA-Net} proposes a bundle adjustment layer to predict the damping factor of the Levenberg-Marquardt algorithm~\cite{numerical_optimization} and calculates the update. To further reduce the optimization space, BA-Net also parameterizes the depth map as a linear combination of $128$ single-view predicted basis maps. Utilizing the information of the whole image, the above methods generate robust camera poses and smooth depth maps. Although these methods achieve impressive results compared with traditional methods, they need multiple iterations (e.g. 15 iterations in LS-Net and BA-Net) to converge, and most methods  (e.g. LS-Net and DeMoN) estimate the depth map using only one target frame.
 
\begin{figure}[t]
   
\begin{center}
       
\includegraphics[width=1.0\linewidth]{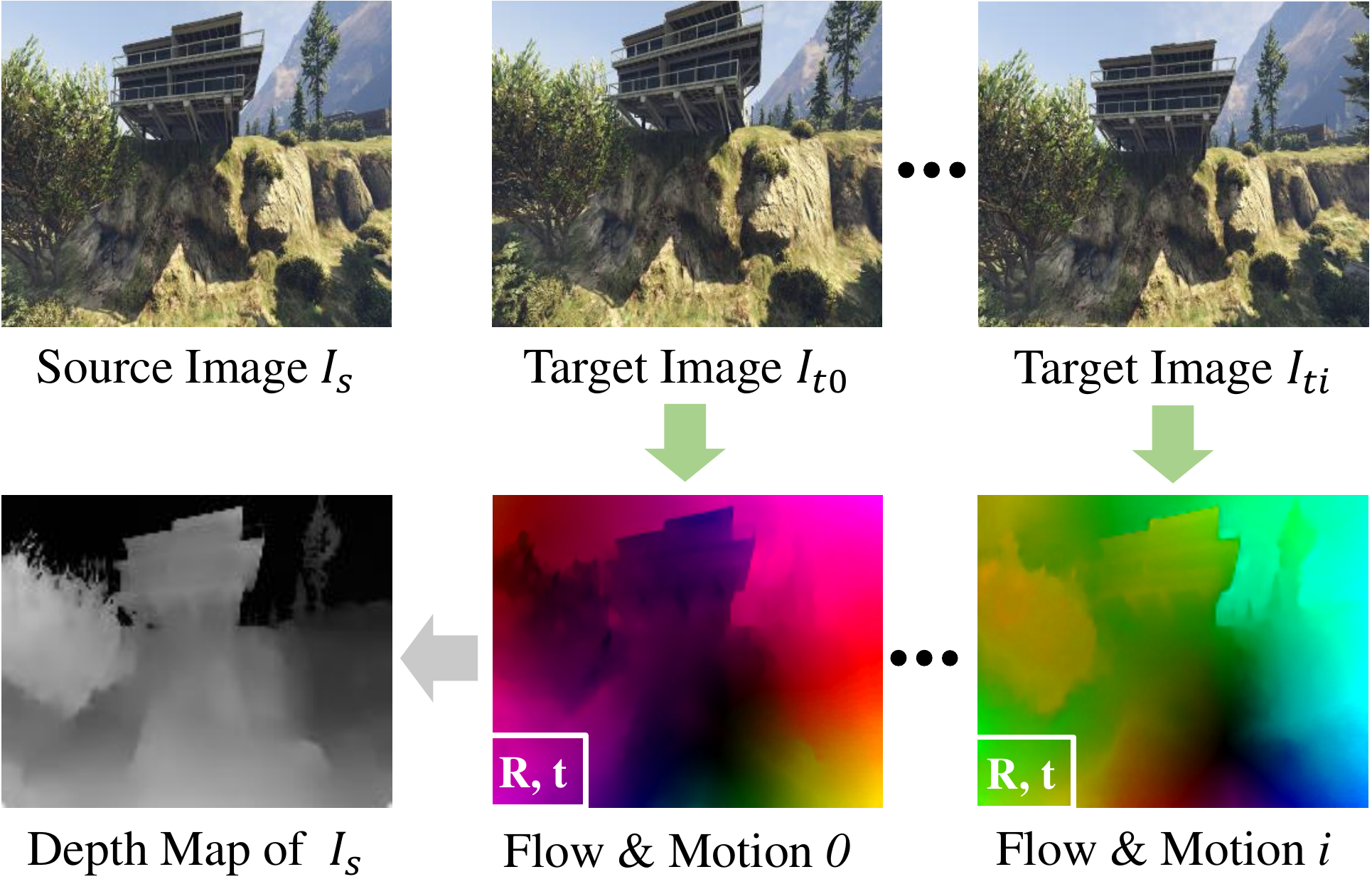}
   
\end{center}
   
\vspace{-0.4cm}
   
\caption{Illustration of the proposed method. Given multiple images from a moving monocular camera, the flow-motion network (green in the figure) first estimates the optical flow and camera pose between the source image and each target image. The estimated flow and motion are further fused by the depth network (gray in the figure) to compute the depth map of the source image.}
   
\vspace{-0.6cm}
   
\label{fig:multiview_example}
\end{figure}
 
In this letter, we improve both the efficiency and accuracy of the state of the art by incorporating domain knowledge and further extend the method to fuse multiple depth information. The first contribution of our work is a joint estimation of the optical flow and camera poses. We observe that, in monocular stereo problems, the optical flow between multiview images is caused by the ego-motion of a moving camera in static scenes such that the optical flow is constrained along the epipolar lines. By jointly considering the optical flow and camera poses, the pixel search space can be gradually reduced, improving both the efficiency and accuracy. A novel triangulation layer is proposed to encode the estimated optical flow and camera motion without numerical problems caused by unconstrained camera movements. The encoded information from the triangulation layer is used to estimate the depth map of the source image. In many applications, the source image is observed by multiple target images. Beyond the two-view problem, we further extend the networks to fuse depth information from multiple target frames. The depth information from different image pairs is fused by mean-pooling layers and is then used to predict the depth map. Figure~\ref{fig:multiview_example} shows the workflow of our method estimating the optical flow, camera poses, and depth map given multiple images. By exploiting multiview observations, robust and accurate depth maps can be generated.
 
Training and evaluating learning-based SfM methods requires lots of images with ground truth camera poses and depth maps. Existing datasets, for example, SUN3D~\cite{sun3d} and Scenes11~\cite{demon}, contain either low-quality images from RGB-D cameras or non-photorealistic synthetic images. To train and evaluate our proposed networks, we develop tools that can generate unlimited high-quality photorealistic images with ground truth depth maps and camera poses from the game Grand Theft Auto V (GTA5). For the benefit of the computer vision community, we release the tools and generated datasets as open source.
 
To summarize, the contributions of the letter are the following:
\begin{itemize}
   
\item A network that jointly estimates optical flow and camera poses given two-view images. With the estimated camera poses, the optical flow is constrained on epipolar lines such that the flow can be regularized, and the search space is reduced.
   
\item A novel triangulation layer that encodes the estimated optical flow and camera pose so that the depth network can triangulate the depth of each pixel without numerical problems.
   
\item The depth network is further extended to fuse depth information (e.g. flow and motion) from multiple image pairs. By fusing multiple observations, the depth of the source image can be estimated more accurately and robustly.
   
\item Open source tools to customize unlimited photorealistic synthetic images with different daytime, intrinsic parameters, etc. The extracted images serve as a supplementary dataset to train and evaluate learning-based SfM methods.
\end{itemize}

\section{Related Work}
 
\begin{figure*}[h]
\vspace{0.3cm}
    \begin{center}
        \includegraphics[width=0.9\linewidth]{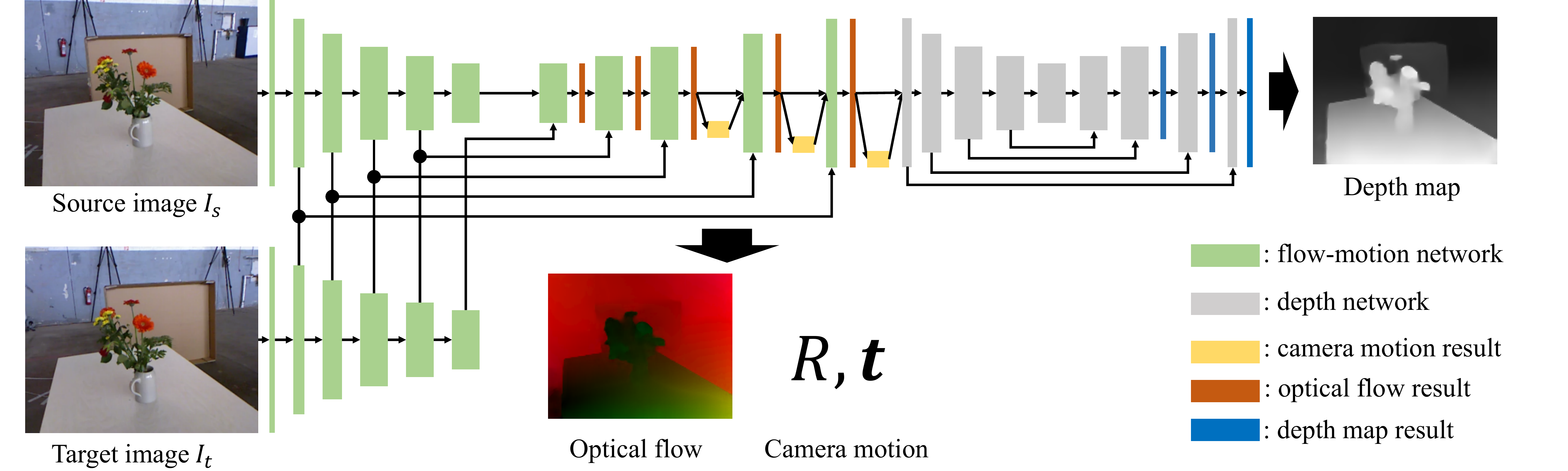}
    \end{center}
\vspace{-0.3cm}
\caption{The architecture of the proposed flow-motion network and depth network. Here, only the two-view architecture is shown for simplicity. The extension to fuse multiple depth information is shown in Figure~\ref{fig:multiview_support} and is discussed in Section~\ref{multiframe_sfm}. The flow-motion network jointly estimates the optical flow and camera poses, and the depth network triangulates the depth of each pixel in the source image. Although both networks are based on simple encoder-decoder architectures, the proposed joint estimation (Section~\ref{flow_motion_net}) and triangulation layer (Section~\ref{depth_net}) enables high-quality and efficient estimation.}
   
\label{fig:architecture}
   
\vspace{-0.5cm}
\end{figure*}
 
In this section, we outline related work using neural networks to estimate the camera poses and depth maps given two or more images.
 
DeMoN~\cite{demon} is a pioneering work that jointly estimates depth maps and camera poses given two-view images. To effectively use the two-view observations, DeMoN adapts FlowNetS~\cite{flownet} to first estimate the optical flow between two images, and then decomposes the flow into camera poses and depth maps. To further improve the quality, DeMoN iteratively refines the optical flow, camera pose and depth map using two encoder-decoder networks, and finally upsamples the depth map into a higher resolution.
 
CodeSLAM~\cite{code_slam} and BA-Net~\cite{BA-Net} parameterize depth maps as compact representations such that both the camera motion and depth map can be solved explicitly by classic optimization methods. CodeSLAM uses an auto-encoder and decoder to represent the depth map as a function of the corresponding image and unknown code. The unknown code can be solved jointly with the camera pose by minimizing the photometric error and geometric error. Benefiting from the flexibility of the classic optimization, CodeSLAM can simultaneously estimate multiple depth maps and camera poses. To make the depth representation suitable for SfM tasks, BA-Net embeds the bundle adjustment as a differentiable layer into the network and the whole process is end-to-end trainable. Unlike CodeSLAM and BA-Net, LS-Net~\cite{ls_net} trains a CNN as a least-square solver to update camera poses and depth values. Starting from initialized depth maps and camera poses, these methods need multiple iterations to converge.
 
 
Many approaches have been proposed to solve multiview stereo or camera tracking using neural networks. Given multiple images with known camera poses and intrinsic calibration, DeepMVS~\cite{DeepMVS} generates cost volumes using learned feature maps and then estimates the disparity map by fusing multiple cost volumes. MVDepthNet~\cite{mvdepthnet}, DPSNet~\cite{DPSNet} and MVSNet~\cite{MVSNet,RMVSNet} solve the same reconstruction problem but differ in the calculation of cost volumes and the structure of networks. On the other hand, given an RGB-D keyframe, DeepTAM~\cite{deeptam} incrementally tracks the pose of a camera using synthetic viewpoints and can further estimate the depth map of the tracked frame.
 
Here, we propose a method that is different from all the monocular stereo methods mentioned above. The major difference is that our method does not iteratively refine the estimation but rather generates results using only one forward pass in the flow-motion network and depth network. The key to the improved efficiency and quality is the joint estimation of both optical flow and camera motion. The high-quality optical flow directly establishes precise dense pixel correspondences between images, enabling accurate depth triangulation. Also, the proposed method can be extended to estimate the depth map of the source image by fusing the information from multiple target images.
 
\section{Network Architecture}
 
As shown in Figure~\ref{fig:architecture}, the proposed method consists of two networks: one flow-motion network and one depth network. Given a source image $I_{s}$ and a target image $I_{t}$ of a static scene, the flow-motion network estimates the optical flow between two images and the relative camera pose in a coarse-to-fine manner. With camera poses estimation, the search space of the optical flow can be gradually reduced along the epipolar line. Moreover, the aperture problem in optical flow can be reduced by the epipolar line constraint. With the estimated optical flow and camera motion, the depth value of each pixel can be directly triangulated. However, the triangulation step is not numerically stable around the epipolar~\cite{multiview_geometry}. Instead, we propose a triangulation layer to encode the information of the estimated optical flow and camera poses. The layer is processed by the depth network to estimate the depth map of $I_s$. The depth network can also be extended to fuse the information from multiple target images. When the source image is observed by multiple target images, the depth map of the source image can be solved by fusing information from \textit{all} source-target pairs.
 
In the following sections, we first explain the design of the flow-motion network, the depth network that process two-frame SfM problems. In Section~\ref{multiframe_sfm}, the depth network is further extended to fuse multiple depth information and estimate the depth map of the source image.
 
\subsection{Flow-Motion Network}\label{flow_motion_net}
 
A number of works~\cite{flownet, flownet2, spy_net, pwc_net} have shown the success of using CNNs to estimate dense optical flow between two images. The proposed flow-motion network shares similar structures to the state-of-the-art PWC-Net~\cite{pwc_net} but is tailored for static scenes and jointly estimates camera poses. 
 
To be robust to lighting and viewing angle changes, input images are converted into L-level feature pyramids using a simple CNN. The feature map at the $i$-th level, $\mathbf{f}^i$, is processed by three simple convolutional layers to generate the next level feature map $\mathbf{f}^{i+1}$ with the size downsampled by $2$. In this work, $L=6$ pyramid levels are used, with $\mathbf{f}^0$ being the original 3-channel image. $\mathbf{f}_s^{i}$ and $\mathbf{f}_t^{i}$ are used to denote the  $i$-th level feature maps of $I_s$ and $I_t$, respectively. 
 
The optical flow $\mathbf{w}$ is estimated from coarse to fine to handle large pixel displacement. At the $i$-th level, the optical flow $\mathbf{w}^{i+1}$ from the \textit{i+1}-th level is firstly bilinear upsampled into $\mathbf{w}^{i+1}_{up}$ as an initialization of $\mathbf{w}^i$. A cost volume $\mathbf{c}^{i}$ is constructed using $\mathbf{f}_s^{i}$ and $\mathbf{f}_t^{i}$. Each element in the cost volume represents the feature similarity between a pixel $\mathbf{x}_s$ in $\mathbf{f}_s^{i}$ and a pixel $\mathbf{x}_t$ in $\mathbf{f}_t^{i}$,
\begin{equation}
\mathbf{c}^{i}(\mathbf{x}_s, \mathbf{x}_t) = \frac{1}{N_i} {(\mathbf{f}_s^{i}(\mathbf{x}_s))}^{T} \mathbf{f}_t^{i}(\mathbf{x}_t),
\end{equation}
and $N_i$ is the feature dimension of $\mathbf{f}_s^{i}$. Due to the coarse-to-fine manner, only a subset of pixels in $\mathbf{f}_t^{i}$ is needed to calculate the cost volume. The cost volume $\mathbf{c}^{i}$, upsampled optical flow $\mathbf{w}^{i+1}_{up}$, and $\mathbf{f}_s^{i}$ are used to predict the optical flow $\mathbf{w}^{i}$ using the DenseNet~\cite{dense_net} structure.

\begin{figure}[t]
\begin{center}
\includegraphics[width=0.9\linewidth]{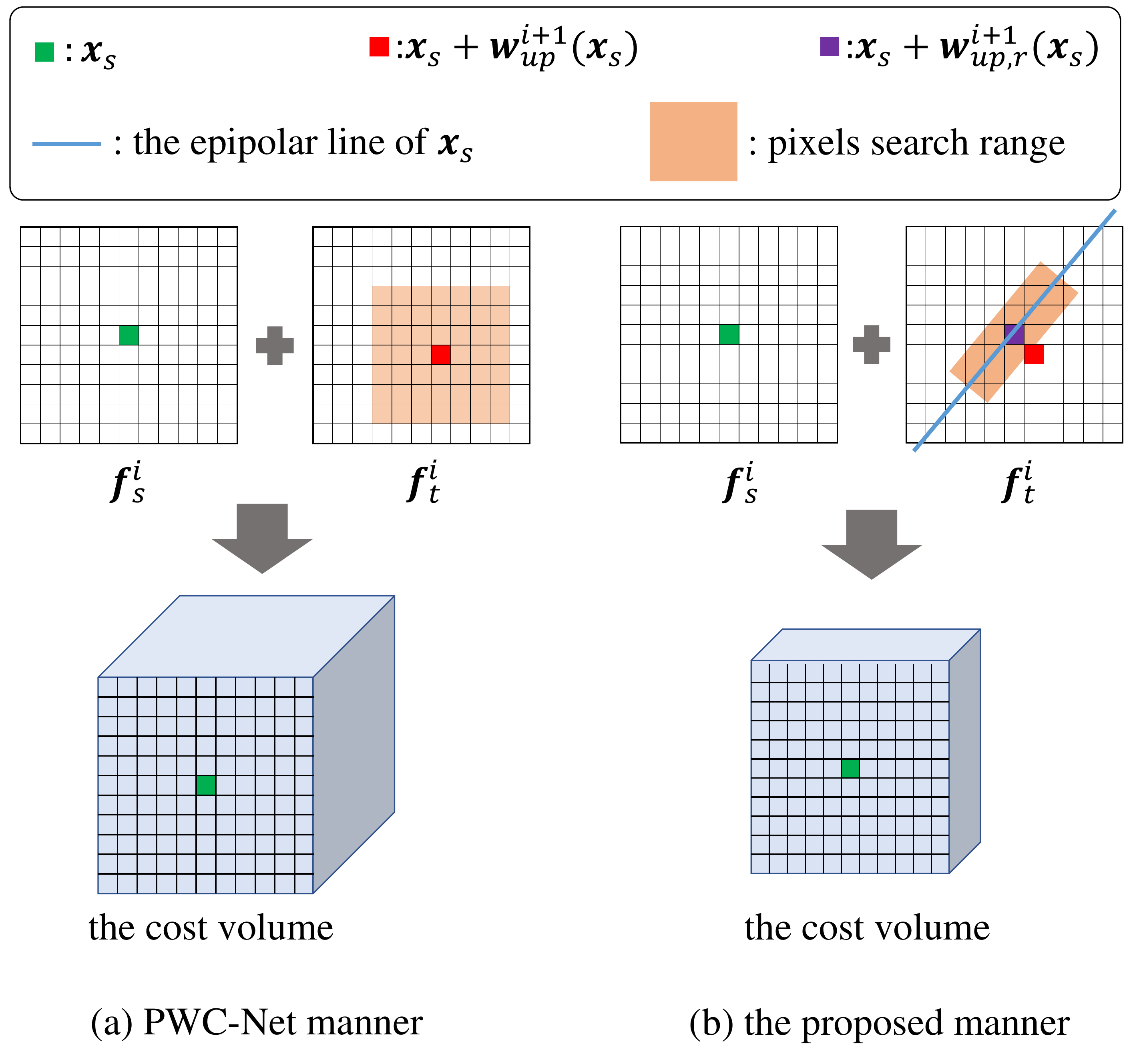}
\end{center}
\vspace{-0.5cm}
\caption{Differences between the cost volume computation in PWC-Net (left) and the proposed flow-motion network (right). For each pixel $\mathbf{x}_s$ in $\mathbf{f}_s^i$, PWC-Net matches a fixed set of pixels (colored in orange) around $\mathbf{x}_s + \mathbf{w}_{up}^{i+1}(\mathbf{x}_s)$ to generate the cost volume. On the other hand, the proposed flow-motion network first regularizes the initial flow $\mathbf{w}_{up}^{i+1}(\mathbf{x}_s)$ into $\mathbf{w}_{up,r}^{i+1}(\mathbf{x}_s)$ and matches pixels around the epipolar line.}
\label{fig:cost_volume}
\vspace{-0.5 cm}
\end{figure}
 
The above cost volume construction and optical flow estimation are repeated from coarse to fine until the optical flow of the desired resolution is estimated. In this work, we adapt and improve the above processes by incorporating the static scene prior and jointly estimating the camera pose.
 
In different pyramid levels, several convolutional layers and linear layers are used to predict the pose of the source frame with respect to the target frame. The pose consists of a rotation matrix $R$ and a translation vector $\mathbf{t}$. With the estimated camera motion and calibrated intrinsic $K$, the flow vector of each pixel can be regularized along the corresponding epipolar line and the search space of pixels in the cost volume can be narrowed down.
 
In static environments, pixel $\mathbf{x}$ in the source image and its optical flow vector $\mathbf{w}(\mathbf{x})$ to the target image have the following relationship,
\begin{equation}
\begin{bmatrix} \mathbf{x}  + \mathbf{w}(\mathbf{x}) \\ 1 \end{bmatrix}^T F \begin{bmatrix} \mathbf{x} \\ 1 \end{bmatrix} = 0,
\end{equation}
where $F = K^{-T} \mathbf{t}_\times R K^{-1}$ is the fundamental matrix. With the estimated camera pose, the upsampled optical flow of each pixel $\mathbf{w}^{i+1}_{up}(\mathbf{x})$ can be regularized by projecting the corresponding point to the epipolar line,
\begin{equation}
\mathbf{w}^{i+1}_{up,r}(\mathbf{x}) = \frac{1}{e_x^2+e_y^2}
\begin{bmatrix} 
\scalebox{0.9}{$x' e_y^2 - y' e_x e_y - e_x e_z$} \\[0.15cm]
\scalebox{0.9}{$y' e_x^2 - x' e_x e_y - e_y e_z$}
\end{bmatrix} - \mathbf{x},
\end{equation}
where \scalebox{0.9}{$\small[e_x, e_y, e_z]^T = F [\mathbf{x}, 1]^T$} and \scalebox{0.9}{$[x', y']^T = \mathbf{x} + \mathbf{w}^{i+1}_{up}(\mathbf{x})$}.
 
Since the corresponding pixels are constrained on epipolar lines, it is not necessary to match pixels far from the lines. Also, the aperture problem, where the pixel correspondences cannot be determined due to the ambiguous matchings, can be reduced by incorporating the epipolar line constraint. However, the epipolar lines, which are determined by the estimated camera poses, may not be accurate enough to rule out all pixels off the lines. Here, we gradually decrease the search space from coarse pyramid levels to fine levels. In the $i$-th level, the matching pixels of pixel $\mathbf{x}_s$ is parameterized as 
\begin{equation}
\scalebox{0.9}{$
\begin{aligned}
\mathbf{x}_t \in \{ & \mathbf{x}_s + \mathbf{w}^{i+1}_{up,r}(\mathbf{x}_s) + \frac{h(e_y,-e_x)^T + v(e_x,e_y)^T}{e_x^2 + e_y^2} \mid \\ 
& h \in [-h_{max}^i,h_{max}^i], v \in [-v_{max}^i, v_{max}^i]\},
\end{aligned}$}
\end{equation}
where $h_{max}^i$ denotes the search range along the epipolar lines and $v_{max}^i$ is the search range vertical to the lines. In total, $(2h_{max}^i+1)(2v_{max}^i+1)$ pixels are matched for each pixel at the $i$-th level.
 
Figure~\ref{fig:cost_volume} illustrates the difference between the cost volume computation in PWC-Net and the proposed flow-motion net. With the static scene prior and the estimated motion, the estimated optical flow can be regularized, and the size of the cost volume is reduced, leading to efficient estimation.
 
\begin{figure}[t]
\begin{center}
\includegraphics[width=0.9\linewidth]{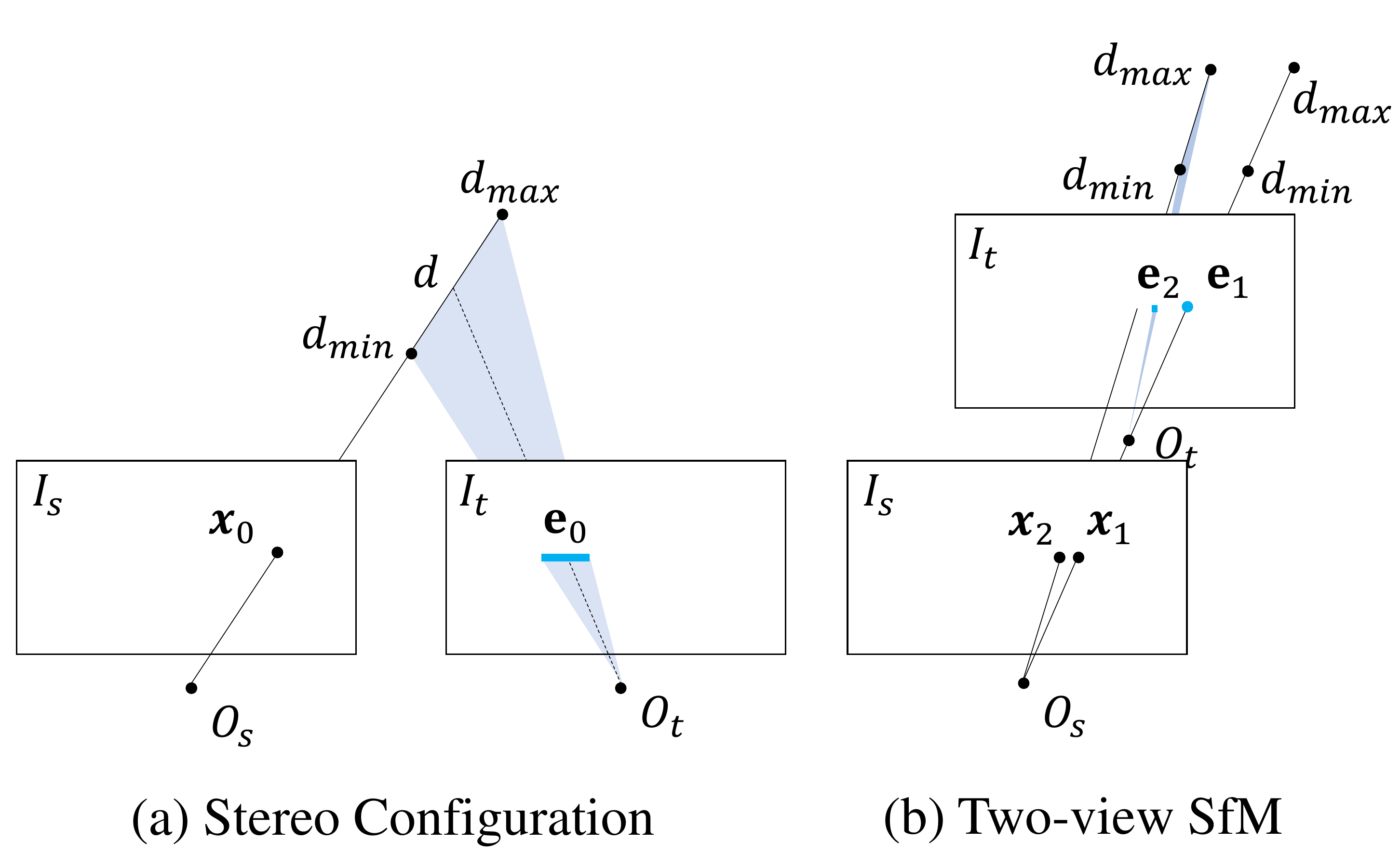}
\end{center}
\vspace{-0.5cm}
\caption{Example to show the numerical stability in triangulation. $O_s$ and $O_t$ are the optical centers of $I_s$ and $I_t$, respectively. $d_{max}$ and $d_{min}$ are the maximum and minimum depth of the scene. $\mathbf{e}_i$ is the corresponding epipolar line of pixel $\mathbf{x}_i$. (a) In stereo configurations, the depth can be reliably calculated by finding the corresponding point on $\mathbf{e}_0$. (b) In unconstrained monocular stereo problems, the epipolar line $\mathbf{e}_1$ of $\mathbf{x}_1$ (\textit{the epipolar point}) degenerates into a point, thus the depth is unobservable. For pixels near the epipolar point, such as $\mathbf{x}_2$, the epipolar line $\mathbf{e}_2$ is very short, and the result is noise-prone.}
\label{fig:stereo_vs_mono}
\vspace{-0.5cm}
\end{figure}
 
\subsection{Depth Network}\label{depth_net}
 
Given the estimated optical flow $\mathbf{x}$ and camera pose $R$, $\mathbf{t}$, the pixel depth $d$ can be easily triangulated by solving,
\begin{equation} \label{triangulate}
\mathbf{w}(\mathbf{x}) + \mathbf{x} = \lambda(KRK^{-1}[\mathbf{x}, 1]^T \cdot d + K\mathbf{t}),
\end{equation}
where $\lambda([x,y,z]^T) = [x/z,y/z]^T$ is the dehomogenization function. However, two drawbacks exist in this triangulation step. First, the depth is solved independently for each pixel, thus the overall smoothness and scene priors are ignored. Second, pixels around the epipolar (the projection of the target frame's optical center on the source image) cannot be triangulated reliably. Figure~\ref{fig:stereo_vs_mono} illustrates the potential numerical issues in different camera motions.
 
To solve the above problems, DeMoN uses networks to refine the triangulated depth maps (with affected depth set to 0). Here, instead of refining the triangulated depth maps, we propose an eight-channel layer that encodes all the information for triangulation. The layer is called triangulation layer $\mathbf{tri}$, and for each pixel $\mathbf{x}$,
\begin{equation}
\mathbf{tri}(\mathbf{x}) = [\mathbf{w}(\mathbf{x}) + \mathbf{x}, KRK^{-1}[\mathbf{x}, 1]^T, K\mathbf{t}]^T.
\end{equation}
 
The depth network is an encoder-decoder network that takes the triangulation layer $\mathbf{tri}$, source image $I_s$, estimated optical flow $\mathbf{w}$ and the last layer of the flow-motion network as input to estimate the depth map of the source image.
 
\subsection{Multiview Depth Fusion}\label{multiframe_sfm}
 
\begin{figure}[t]
\begin{center}
\includegraphics[width=0.9\linewidth]{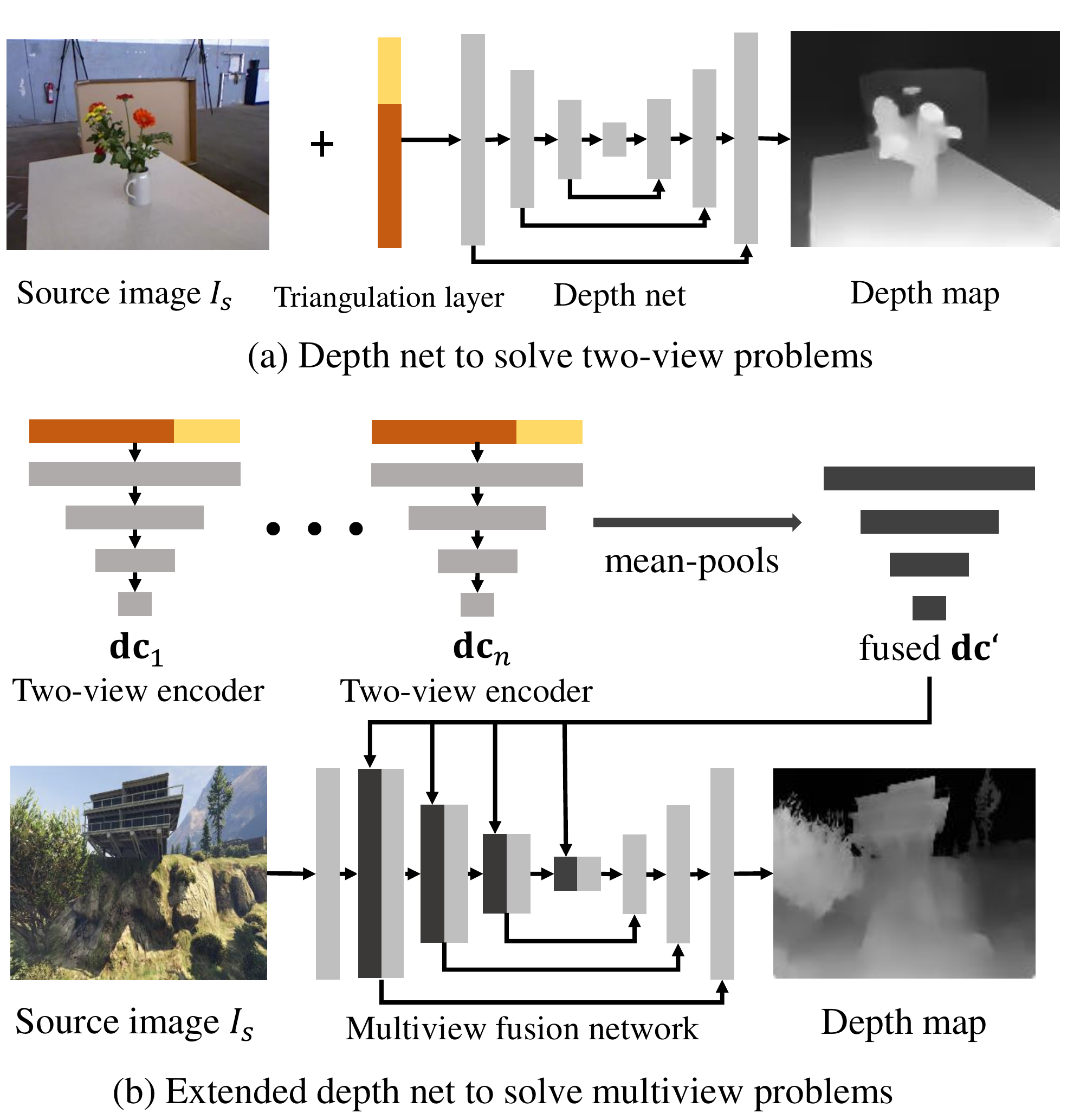}
\end{center}
\vspace{-0.3cm}
\caption{Extending the depth net to fuse multiple depth information. (a) Two-view depth estimation network. (b) Multiple depth fusion extension. The \textit{two-view encoder network} encodes the depth information of each image pair into depth codes $\mathbf{dc}_i$. Multiple codes are pooled into $\mathbf{dc}'$ and the \textit{multiview fusion network} takes $\mathbf{dc}'$ to estimate the depth map.}
\label{fig:multiview_support}
\vspace{-0.5cm}
\end{figure}
 
In real-world applications (e.g. robot navigation), the depth of the source image can be solved by multiple target images. Here, we extend the proposed two-view monocular stereo networks to fuse multiview information. Compared with two-view image pairs, multiview images bring more information about the environment structure, thus the fused depth maps can be more robust and accurate. However, fusing depth information from multiview images is non-trivial due to the arbitrary number of image pairs and different depth scales. Different from CodeSLAM, which fuses the information by optimization methods, we propose to fuse the multiview information by a learned network.
 
Figure~\ref{fig:multiview_support} shows how the two-view depth net is extended. The two-view depth net introduced in Sec.~\ref{depth_net} is divided into two parts: two-view encoder and multiview fusion. The first part independently encodes the triangulation layer $\mathbf{tri}$ of each image pair into multi-resolution depth codes $\mathbf{dc}$. Depth codes from multiple image pairs, \{$\mathbf{dc}_0$, ..., $\mathbf{dc}_{N-1}$\}, are fused by mean-pooling layers. The fused code of each pixel $\mathbf{dc}'(\mathbf{x})$ is calculated as,
\begin{equation}
\mathbf{dc}'(\mathbf{x}) = \frac{1}{N}\sum^{N-1}_{i=0}\mathbf{dc}_i(\mathbf{x}).
\end{equation}

Using pooling layers to fuse information has been used in many multiview stereo works (e.g., DeepMVS~\cite{DeepMVS}). Different from these works, we use multiple pooling layers to fuse the depth codes at different resolutions such that both the global information and fine details are preserved. The fusion network takes the fused depth code $\mathbf{dc}'$ and the source image $I_s$ to estimate the corresponding depth map.
 
\section{Network Details}
 
\subsection{Optical Flow and Camera Motion}
 
The search space of the cost volume calculation is reduced gradually from coarse to fine. The flow-motion network estimates the optical flow from level $5$ to level $1$. From the $5$-th level to the $1$-st level, the search steps $h_{max}$ and $v_{max}$ are set to $\{4,4,4,4,3\}$ and $\{4,4,4,2,1\}$, respectively. In the $1$-st level, only $21$ pixels are matched ($81$ pixels are used in PWC-Net). The optical flow loss is defined as,
\begin{equation}
L_{flow} = \sum_{l=1}^{5} \sum_{\mathbf{x}} { \lVert \mathbf{w}^l(\mathbf{x}) - \hat{\mathbf{w}}^l(\mathbf{x}) \rVert }_2,
\end{equation}
where $\hat{\mathbf{w}}^l$ is the corresponding ground truth optical flow at the $l$-th level.
 
The camera rotation $\mathbf{r}$ is parameterized as the three-dimension rotation vector: $\mathbf{r} = \theta\mathbf{v}$, where $\theta$ is the rotation angle and $\mathbf{v}$ is the rotation axis. Similar to DeMoN, camera translation $\mathbf{t}$ is normalized as a unit vector due to the unobservable scale. Since the optical flow on coarse resolutions cannot provide accurate pixel correspondences, the camera motion is estimated from level $3$ to level $1$. With the ground truth camera motion $\hat{\mathbf{r}}$ and $\hat{\mathbf{t}}$, the motion loss is,
\begin{equation}
L_{motion} = \sum_{l=1}^{3} { \lVert \mathbf{r}^l -\hat{\mathbf{r}} \rVert }_2 + \sum_{l=1}^{3} { \lVert \mathbf{t}^l -\hat{\mathbf{t}} \rVert }_2.
\end{equation}
 
\subsection{Depth Estimation}
 
Multiple depth maps are estimated by the depth network at different resolutions (from level $3$ to level $1$). We adopt the depth parameterization from Eigen et al.~\cite{mono_depth_2014} that the output of the network is the log depth: $\text{log}(d) \in R$. Due to the scale ambiguity in SfM problems, the scale-invariant depth error for each pixel $\mathbf{x}$ is calculated as,
\begin{equation}
d_{e}^l(\mathbf{x}) = \text{log}(d^l)(\mathbf{x}) + \alpha^l - \text{log}(\hat{d^l})(\mathbf{x})
\end{equation}
where $\hat{d}$ is the ground truth depth map, and \scalebox{0.9}{$\alpha^l = \frac{1}{N} \sum_{\mathbf{x}} \text{log}(\hat{d^l})(\mathbf{x}) - \text{log}(d^l)(\mathbf{x})$} scales the estimated depth maps. Both the depth error $L_{d}$ and gradient error $L_{g}$ are calculated to train the triangulation network,
\begin{equation}
    \scalebox{0.9}{$
L_{d} = \sum_{l=1}^{3} \sum_{\mathbf{x}} { \lVert d_{e}^l(\mathbf{x}) \rVert }_{berHu},
$}
\end{equation}
\begin{equation}
\scalebox{0.9}{$
L_{g} = \sum_{l=1}^{3} \sum_{\mathbf{x}} \left| \nabla_x d_{e}^l(\mathbf{x}) \right| + \left| \nabla_y d_{e}^l(\mathbf{x}) \right|,
$}
\end{equation}
where ${ \lVert \cdot \rVert }_{berHu}$ is the reverse Huber~\cite{fcrn_mono_depth, berhu_loss}:
\begin{equation}
\scalebox{0.9}{$
    { \lVert x \rVert }_{berHu} =
\begin{cases} 
\left| x \right| & \text{if}\left| x \right| \leq 1\\
x^2 & \text{if}\left| x \right| > 1
\end{cases}$}.
\end{equation}
Using the berHu norm, large depth errors are punished by the L2 norm and small depth errors can also be effectively optimized by the L1 norm.

\section{Datasets}
\subsection{DeMoN Dataset}
 
DeMoN proposes a collection of datasets to train and evaluate deep networks. The dataset contains images from multiple sources, such as RGB-D cameras~\cite{sun3d,sturm12iros}, multiview SfM results~\cite{sfm_revist,schonberger_2016, mve_2014,ummenhofer_2015}, and synthetic images~\cite{demon}. In total, the DeMoN dataset contains $57$k image pairs for training and $354$ pairs for testing.
 
Although the DeMoN dataset has been widely used in previous works~\cite{demon, ls_net, BA-Net}, it contains several limitations. First, depth maps from RGB-D cameras are not synchronized with the color images and only provides less than $10$ meters depth measurements. Second, most of the camera poses of the real-world images are calculated by optimization-based methods which can be affected by image noises or outlier features. Lastly, the rendered synthetic images in the dataset are not photorealistic. All these aspects limit the performance of the trained networks.
 
\subsection{GTA-SfM Dataset}
 
\begin{figure}[t]
\begin{center}
\vspace{0.3cm}
\includegraphics[width=0.85\linewidth]{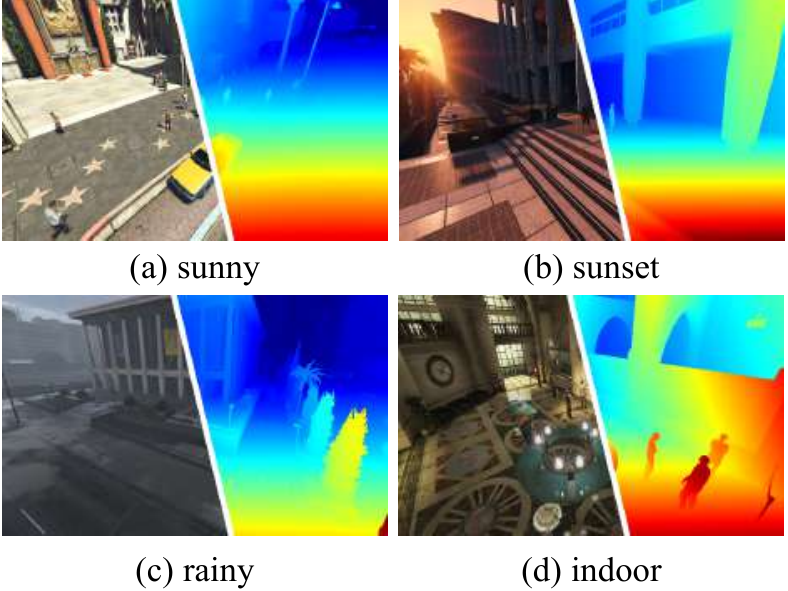}
\end{center}
\vspace{-0.3cm}
\caption{Samples from the GTA-SfM dataset including different weather, time, and scenes. The flexibility to change the environment and camera settings improves the usability of the dataset in deep learning research.}
\label{fig:extracted}
\vspace{-0.5cm}
\end{figure}
 
To overcome the limitations in the DeMoN dataset, we propose the GTA-SfM dataset as a supplement. The dataset is rendered from GTA-V, an open-world game with large-scale city models. Thanks to the active community, we develop tools to extract unlimited photorealistic images with depth maps and camera poses. The extracted depth maps provide depth measurements for all objects in the images, including fine structures or reflective surfaces. We extracted $71k$ pairs of images for training and $5k$ pairs for testing. Training and testing dataset do not share common scenes. Different from the DeMoN dataset, one source image can have multiple target images, thus the multiview depth fusion can be tested.
 
A similar dataset, MVS-SYNTH, is released by DeepMVS~\cite{DeepMVS} using graphics debugging tools. Compared with MVS-SYNTH, GTA-SfM tools can freely set the camera FOV, weather, and daytime such that the dataset diversity and usability are improved. Also, the camera trajectory is manually annotated that cameras move with large translation and rotation. Figure~\ref{fig:extracted} shows samples from the proposed dataset.
 
\section{Experiments}
 
\begin{figure*}[t]
\begin{center}
\includegraphics[width=1.0\linewidth]{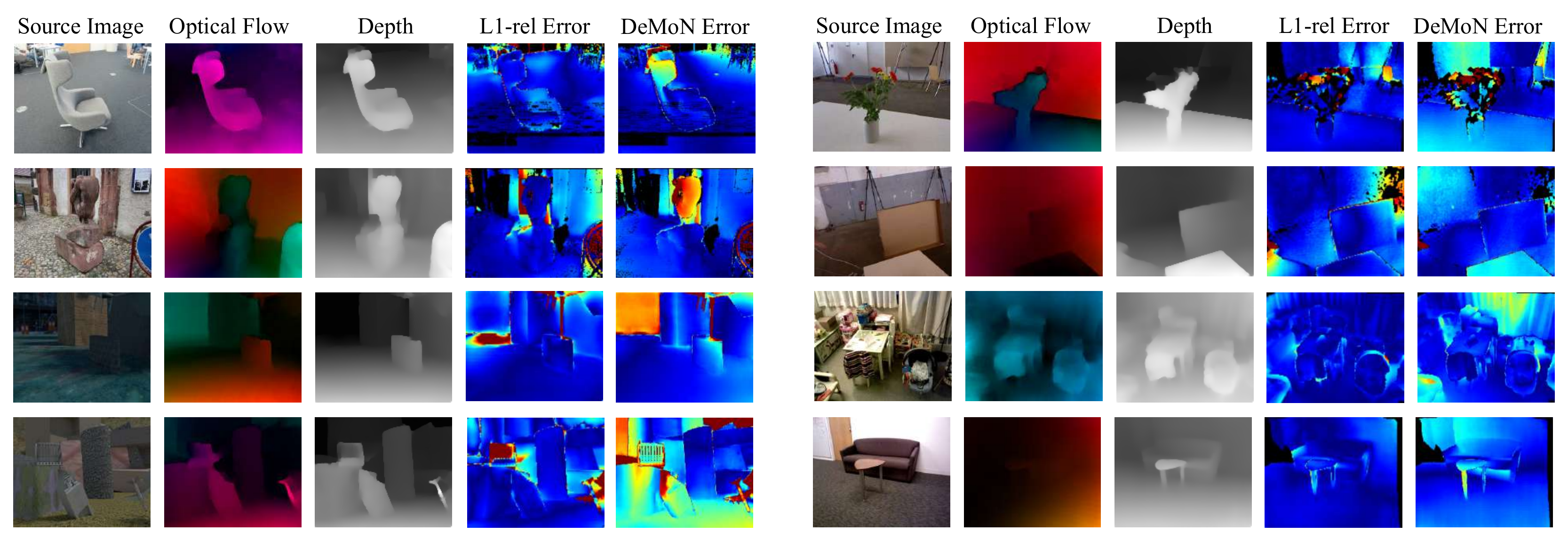}
\end{center}
\vspace{-0.4cm}
\caption{Qualitative results on the DeMoN database. From left to right: source image, estimated optical flow, estimated depth map, L1-rel error map, and L1-rel error of DeMoN estimated depth map. The error map is JET color coded. As shown, our method generates high-quality optical flow and depth maps. With the proposed triangulation layer, the depth maps have less L1-rel errors.}
\label{fig:quality_compare}
\vspace{-0.4cm}
\end{figure*}
 
In this section, we extensively evaluate the performance of the proposed flow-motion network and depth network. We first compare the proposed network with the previous works~\cite{demon,ls_net,BA-Net} on two-view image pairs using the DeMoN dataset. Then, the depth fusion performance is evaluated using the proposed GTA-SfM dataset. The effectiveness of the proposed flow-motion joint estimation and the triangulation layer $\mathbf{tri}$ is also demonstrated in the ablation study. We further demonstrate the generalization ability of the method using real-world images and Google Earth images.
 
\subsection{Evaluation Metrics}
Different metrics are used to evaluate the estimated camera motion and depth maps. We follow the evaluation method used in DeMoN. The rotation error is defined by the relative angle between the estimated camera rotation and the ground truth rotation. Due to the scale ambiguity in SfM problems, the translation error is defined by the angle between normalized translation vectors. For the depth evaluation, estimated depth $d$ is first optimally scaled~\cite{demon}, then three depth metrics are calculated,
\begin{equation}
\scalebox{0.9}{$\text{L1-inv}(d,\hat{d}) = \frac{1}{N} \sum_\mathbf{x} \left| 1 / d(\mathbf{x}) - 1 / \hat{d}(\mathbf{x}) \right|$},
\end{equation}
\begin{equation}
\scalebox{0.9}{$\text{sc-inv}(d,\hat{d}) = \sqrt{\frac{1}{N}\sum_\mathbf{x}z(\mathbf{x})^2 - \frac{1}{N^2}(\sum_\mathbf{x}z(\mathbf{x}))^2}$},
\end{equation}
\begin{equation}
\scalebox{0.9}{$\text{L1-rel}(d,\hat{d}) = \frac{1}{N} \sum_\mathbf{x}  \left| d(\mathbf{x}) - \hat{d}(\mathbf{x}) \right|  / \hat{d}(\mathbf{x})$},
\end{equation}
where \scalebox{0.8}{$z(\mathbf{x}) = \text{log}(d(\mathbf{x})) - \text{log}(\hat{d}(\mathbf{x}))$}, and $N$ is the pixel number.

\subsection{Two-view Evaluation}
 
\begin{table}[t]
\centering
\vspace{0.3cm}
\caption{Comparison on Two-view Problems}
\label{two_view_compare}
\includegraphics[width=0.9\linewidth]{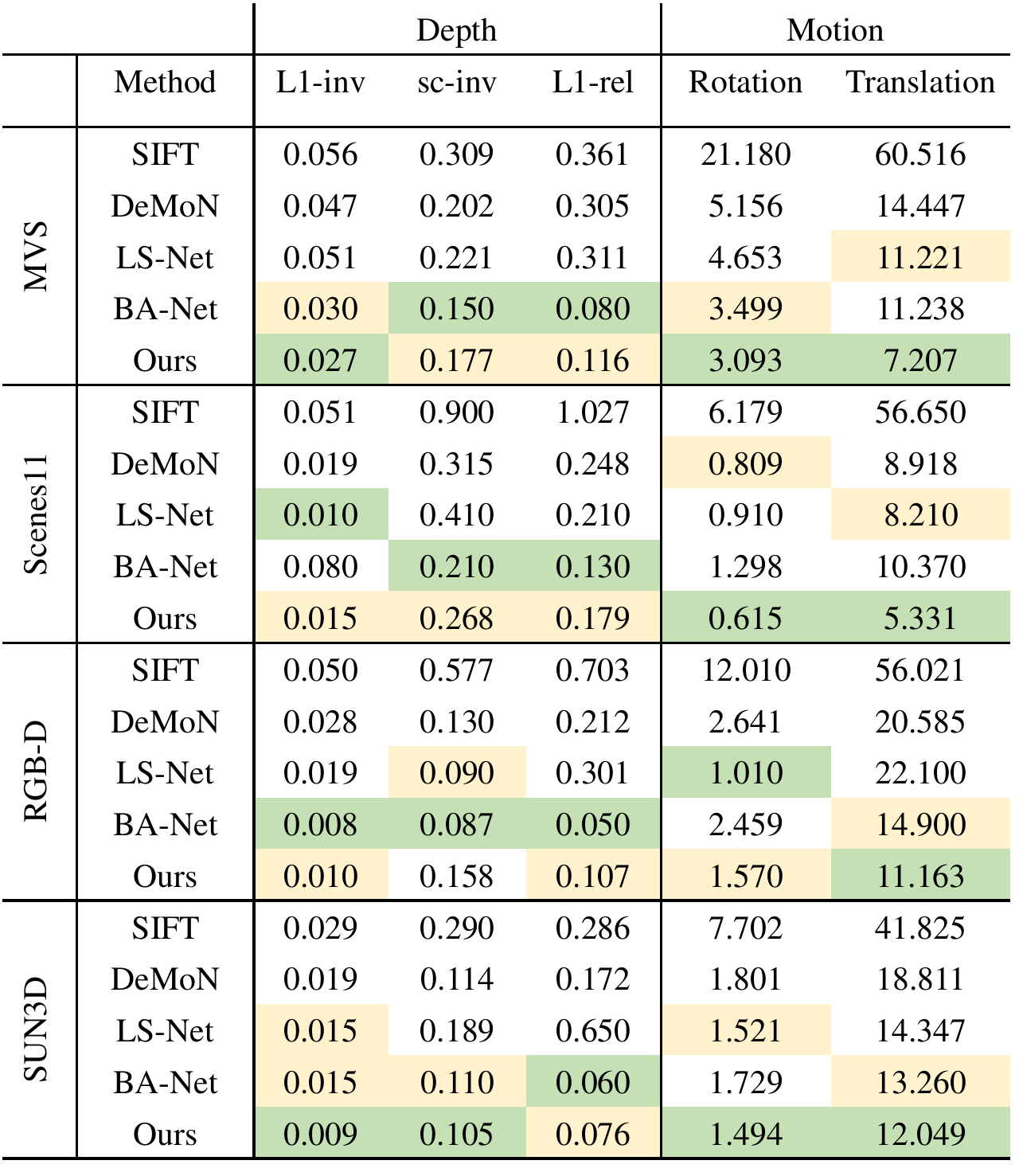}
\vspace{-0.7cm}
\end{table}
 
We train the flow-motion network and the depth network using \textit{only} the DeMoN dataset for a fair comparison. Note that DeMoN is trained with a larger dataset including other synthetic images. Images are resized to $320\times256$ in the experiments. The flow-motion network was trained for $750k$ steps with the Adam optimizer~\cite{adam}. With the trained flow-motion network, the depth network is trained for $260k$ steps. According to the model size, the mini-batch size is set to $16$ for the flow-motion network and $24$ for the triangulation network. Both learning-based methods (DeMoN, LS-Net, and BA-Net) and a classic method are compared in the experiment. The classic method is proposed and evaluated in DeMoN that solves camera poses by the normalized 8-point algorithm~\cite{8_point} (using SIFT features) followed by a reprojection error minimization. The depth maps are estimated using plane sweep stereo and semi-global matching~\cite{hirschmuller2008stereo}.
 
Table~\ref{two_view_compare} and Figure~\ref{fig:quality_compare} show the results of both depth and motion comparison. Due to the flow-motion joint estimation, the proposed method achieves the best camera motion estimation in most of the cases. The proposed depth network also achieves consistently better performance compared with DeMoN. Compared with BA-Net which iteratively refines the results ($95ms$ in total), our method generates consistently better camera poses, and competitive depth maps without any iterations ($42ms$ in total). As shown in Figure~\ref{fig:quality_compare}, due to the triangulation layer that encodes the geometric information, both near and distant objects are reconstructed accurately.

\subsection{Depth Fusion Evaluation}
 
Since the DeMoN dataset only provides two-view image pairs, we use the proposed GTA-SfM dataset to train and evaluate the multiview depth fusion performance. We first train the flow-motion network using two-view image pairs for $210k$ steps and then train the extended multiview fusion network for $130k$ steps. 
 
\begin{figure}[h]
\begin{center}
\includegraphics[width=0.90\linewidth]{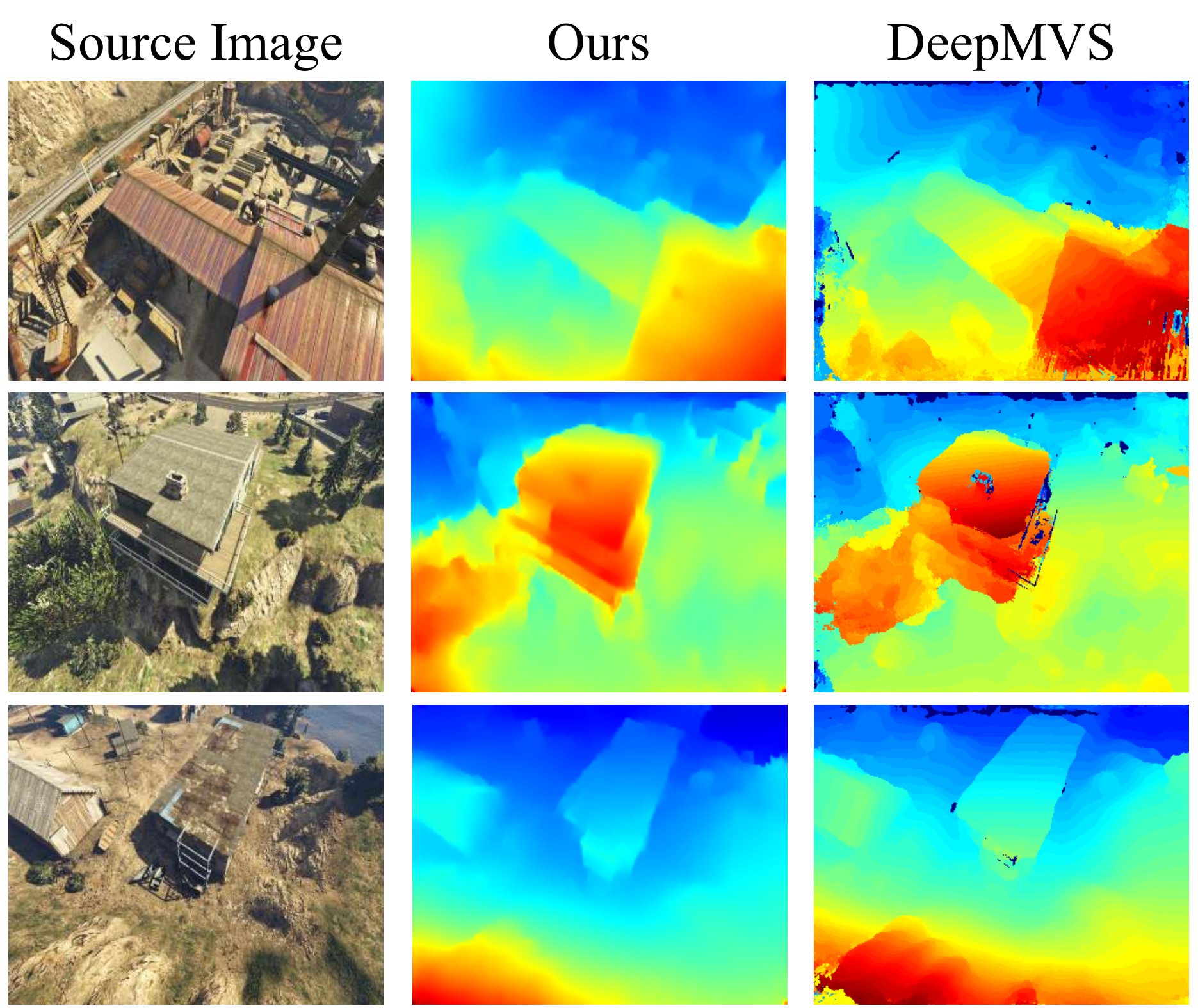}
\end{center}
\vspace{-0.2cm}
\caption{Quality comparison of generated depth maps by the proposed method and DeepMVS. Each source image is observed by 6 target images, and DeepMVS is provided with ground truth camera poses.}
\vspace{-0.2cm}
\label{fig:deepmvs}
\end{figure}

We first evaluate the quality of estimated depth maps using different numbers of target images. We also compare the depth net with DeepMVS~\cite{DeepMVS} which is also trained using images from GTA5. DeepMVS takes ground truth camera poses as input. For each number of target images, we \textit{randomly} sample $300$ pairs and compute the mean depth error. Table~\ref{gta_sfm_depth_map} shows the depth quality given different numbers of target images. Clearly, the depth quality improves when more images are observed, which shows the effectiveness of the multiview fusion and matches the experience from classic SfM methods. We also visualize estimated depth maps for qualitative comparison in Figure~\ref{fig:deepmvs}. Our method estimates smooth and detailed depth maps and DeepMVS estimates discrete depth maps with outliers.
 
\begin{table}[h]
\centering
\caption{Depth map error on GTA-SfM dataset.}
\label{gta_sfm_depth_map}
\vspace{-0.2cm}
    \scalebox{0.8}{
\begin{tabular}{|c|cc|cc|cc|}
\hline
& \multicolumn{6}{c|}{Depth Error}
\\ \hline
\multirow{2}{*}{View Num} & \multicolumn{2}{c|}{L1-inv (1e-3)} & \multicolumn{2}{c|}{sc-inv} & \multicolumn{2}{c|}{L1-rel} \\ \cline{2-7} 
                                    & Ours           & DeepMVS           & Ours        & DeepMVS      & Ours        & DeepMVS       \\ \hline
       
2                         & 6.19           & 16.6              & 0.213       & 0.526        & 0.145       & 0.766         \\
       
3                         & 6.07           & 15.6              & 0.207       & 0.496        & 0.137       & 0.753         \\
        4                         & 5.36           & 15.1              & 0.192       & 0.475        & 0.124       & 0.735         \\
       
5                         & 5.68           & 14.8              & 0.192       & 0.465        & 0.123       & 0.723         \\
       
6                         & 4.86           & 14.8              & 0.181       & 0.464        & 0.114       & 0.729         \\ \hline
       
\end{tabular}}
\vspace{-0.5cm}
\end{table}
 
\subsection{Ablation Study}
 
Here, we study the effectiveness of the contributions: the flow-motion joint estimation and the triangulation layer.
 
\textbf{Joint Estimation} To evaluate the importance of the epipolar line constraint and search space reduce, we remove the camera pose estimation in middle levels and the camera motion is estimated using the final flow estimation. Without the epipolar line constraint, $81$ pixels (the same as PWC-Net) are searched at each level. As shown in Table~\ref{joint_study}, the joint estimation improves \textit{both} the optical flow and camera pose estimation.
 
\begin{table}[h]
\centering
\caption{Effectiveness of the joint flow-motion estimation}
\label{joint_study}
\scalebox{1.0}{
\begin{tabular}{c|ccc}
& Rotation Error & Translation Error & Flow Error \\ \hline
original  & 1.879    & 10.307      & 3.472        \\ \hline
w/o joint & 2.043    & 11.703      & 3.567        \\ \hline
\end{tabular}}
\vspace{-0.4cm}
\end{table}
 
\textbf{Triangulation Layer} The triangulation layer is proposed to encode the estimated optical flow and camera motion without any numerical instability. To demonstrate the effectiveness, we replace the triangulation layer with a directly triangulated depth map. Similar to DeMoN~\cite{demon}, NaN values are set to 0. Both the networks are trained with the same flow-motion network as the front-end for $50$ epochs. The comparison is shown in Table~\ref{no_tri}. With the proposed \textbf{tri}, depth network can better exploit estimated optical flow and camera poses.
 
\begin{table}[h]
\centering
\caption{Effectiveness of the triangulation layer.}
\label{no_tri}
\scalebox{1.0}{
\begin{tabular}{c|ccc}
& L1-inv & sc-inv & L1-rel \\ \hline
original & 0.015  & 0.195  & 0.134  \\ \hline
w/o \textbf{tri}  & 0.017  & 0.200 
& 0.140  \\ \hline
\end{tabular}}
\vspace{-0.4cm}
\end{table}
 
\subsection{Generalization Ability}
 
\begin{figure}[t]
\begin{center}
\vspace{0.3cm}
\includegraphics[width=0.95\linewidth]{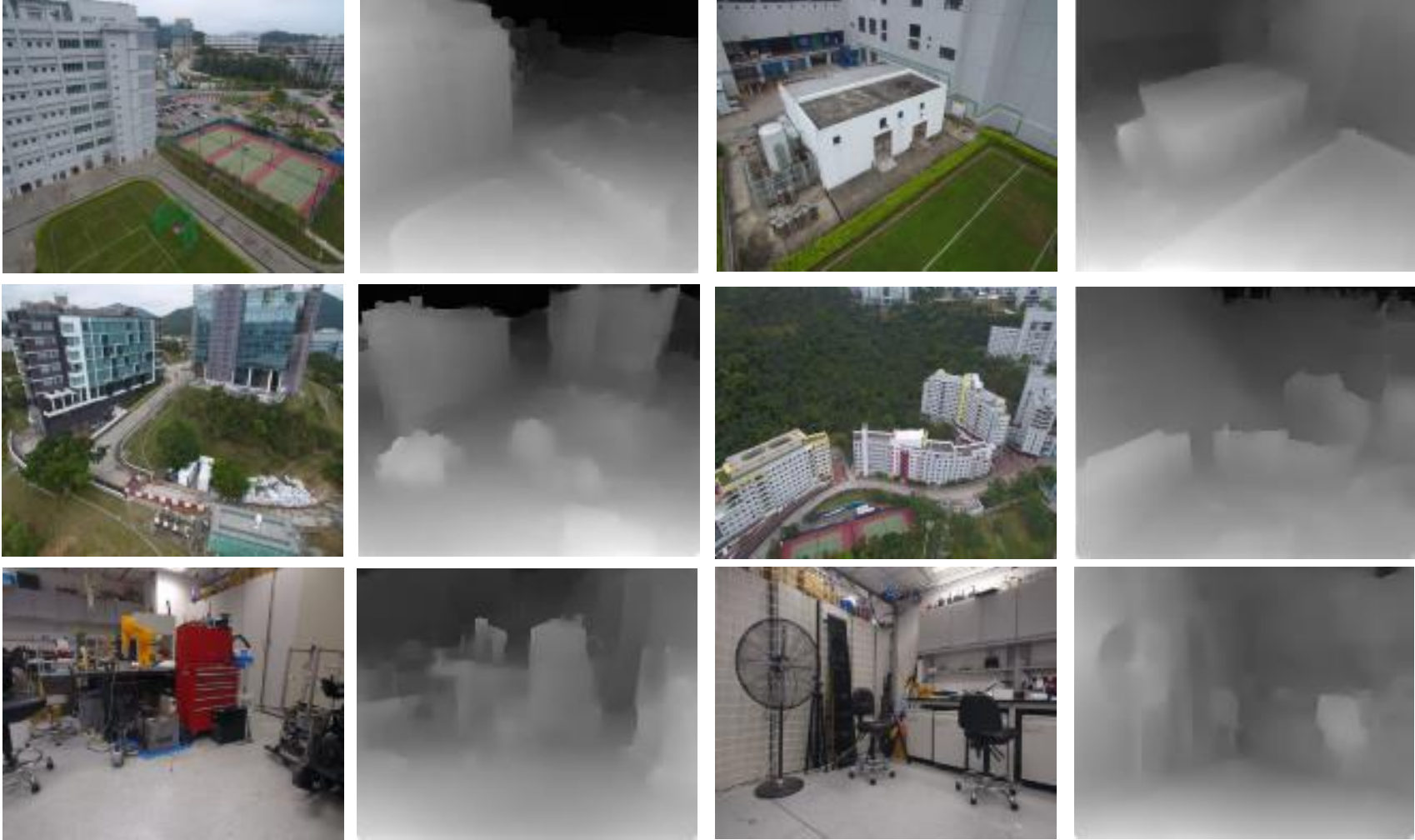}
\end{center}
\vspace{-0.5cm}
\caption{Generate the proposed method to real-world images.}
\label{fig:real_world_dji}
\vspace{-0.3cm}
\end{figure}

\begin{figure}[t]
\begin{center}
\includegraphics[width=0.95\linewidth]{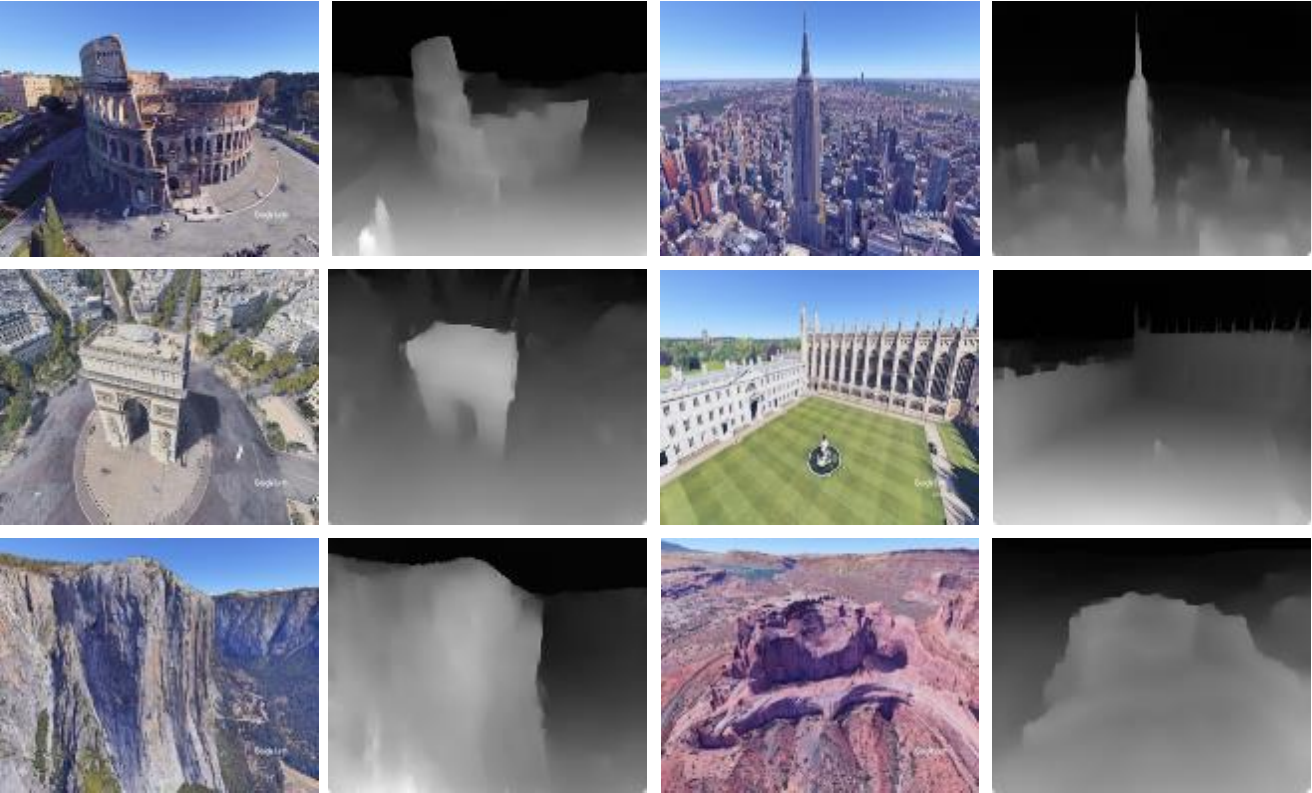}
\end{center}
\vspace{-0.4cm}
\caption{Generate the proposed method to Google Earth images.}
\label{fig:real_world_google}
\vspace{-0.5cm}
\end{figure}

To test the generalization ability of the proposed method, we further use the method to estimate depth maps of images from different sources. Figure~\ref{fig:real_world_dji} shows estimated depth maps of images taken with DJI Phantom 4 (outdoor) or a handheld camera (indoor). Figure~\ref{fig:real_world_google} shows estimated depth maps of images from Google Earth. The depth map of each source image is fused from 5 or 6 target images. Because the proposed method first builds high-quality pixel correspondences and then triangulate the depth of each pixel, it can effectively utilize multiview observations and generalizes well to other images. More details are in the supplementary material.
 
\section{Conclusion and Future Work}
In this letter, we propose a flow-motion network and a depth network that can estimate the camera motion and depth map given multiple motion stereo images. Both the networks are designed carefully to exploit the multiview geometric constraints among optical flow, camera motion and depth maps. We further extend the depth network to fuse multiple depth information into a depth map. To enlarge the available datasets, an open-source tool is proposed to extract unlimited photorealistic images with ground truth camera poses and depth maps. In the future, we plan to further develop the method by incorporating graph networks so that it can simultaneously estimate all camera poses and depth maps given a set of images.
 
{
\bibliographystyle{unsrt}
\bibliography{egbib}
}
 
\end{document}


 
 
 
 
\maketitle

\begin{abstract}
   
This supplementary material provides more details about the proposed method, training, and dataset. More qualitative results of the experiments are provided as a supplement to the main letter. To demonstrate the generalization ability of the network and the effectiveness of the GTA-SfM dataset, the trained network is also directly applied to unseen scenes. Finally, we discuss the advantages and limitations of the proposed method.
\end{abstract}
 
\section{Network Details}
 
In this section, we introduce more details in the flow-motion network and depth network. The depth observability of the epipolar point is also discussed in this section.
 
\subsection{Flow Regularization}
 
With an estimated optical flow $\mathbf{w}$, a regularized flow $\mathbf{w}_r$, is calculated such at the corresponding pixels are constrained on epipolar lines. For each pixel $\mathbf{x}_s$ on the source image, the regularized flow $\mathbf{w}_r(\mathbf{x}_s)$ is calculated by
\begin{equation}
    \scalebox{0.95}{$
    \begin{aligned}
    &\mathop{\arg\min}_{\mathbf{w}_r(\mathbf{x}_s)} && \lVert \mathbf{w}_r(\mathbf{x}_s) - 
    \mathbf{w}(\mathbf{x}_s) \rVert_2 \\
    &\text{subject to} && \begin{bmatrix} \mathbf{x}_s + \mathbf{w}_r(\mathbf{x}_s) \\ 1 \end{bmatrix}^T F \begin{bmatrix} \mathbf{x}_s \\ 1 \end{bmatrix} = 0
    \end{aligned}$}.
\end{equation}
$\mathbf{w}_r(\mathbf{x}_s)$ can be solved as Equation 3 in the main letter.
 
\subsection{Pixel Search Space}
 
$F [\mathbf{x}_s, 1]^T$ is the epipolar line on the target image. For notational simplicity, let $[e_x, e_y, e_z]^T = $ $F [\mathbf{x}_s, 1]^T$. $\mathbf{h} = [e_y, -e_x] / (e_x^2+e_y^2)$ is the normalized vector of the epipolar line direction. One the contrary, $\mathbf{v} = [e_x, e_y] / (e_x^2+e_y^2)$ is the vector perpendicular to $\mathbf{h}$. With $\mathbf{h}$ and $\mathbf{v}$, we define the search space in the target feature map $\mathbf{f}_t$ as Equation 4 in the main letter.
 
\subsection{Depth Observability}
 
The epipolar point in the source image is \scalebox{0.85}{$\mathbf{x}_e = \lambda(-KR^{-1}\mathbf{t})$}. For any pixel depth $d$, the point is projected onto the target image as the same point,
\begin{equation}
\begin{aligned}
\mathbf{x}_t & = \lambda(KRK^{-1}[\mathbf{x}_e, 1]^T \cdot d + K\mathbf{t}) \\
                 & = \lambda(KRK^{-1}[\lambda(-KR^{-1}\mathbf{t}), 1]^T \cdot d + K\mathbf{t}) \\
                 & = \lambda(K\mathbf{t})
\end{aligned},
\end{equation}
thus the depth $d$ is unobservable.
 
For the stereo configuration, $\mathbf{x}_e$ is at infinity so that all pixels in the image can be triangulated. However, in some unconstrained SfM cases, the epipolar point is on the image and cannot be triangulated directly. DeMoN~\cite{demon} uses networks to refine triangulated depth maps with NaN values set to 0. Li \textit{et. al.}~\cite{freeze_people} set pixels around the camera epipolar to zero. We believe the proposed triangulation layer is an alternative solution.
 
 
 
 
 
\section{Training}
 
Since the proposed method decouples the two-view SfM problem into flow-motion estimation and depth triangulation, we train the two networks separately. The flow-motion network is first trained and then is used to train the depth network with weight fixed. Adam optimizer~\cite{adam} is used and the initial learning rate is set to 1e-4. We half the learning rate when the error plateaus. Only color augmentation is used in the training.
 
\section{GTA-SfM dataset}
 
In this letter, we propose a GTA-SfM dataset which is used for the network training. A similar dataset, MVS-Synth, is also rendered in GTA5 environment. In MVS-Synth, cameras usually move randomly with small translations. On the contrary, in the proposed dataset, the trajectory is manually defined that cameras move in large translations and rotations.
 
\begin{figure*}[h]
\vspace{0.5cm}
\begin{center}
\includegraphics[width=1.0\linewidth]{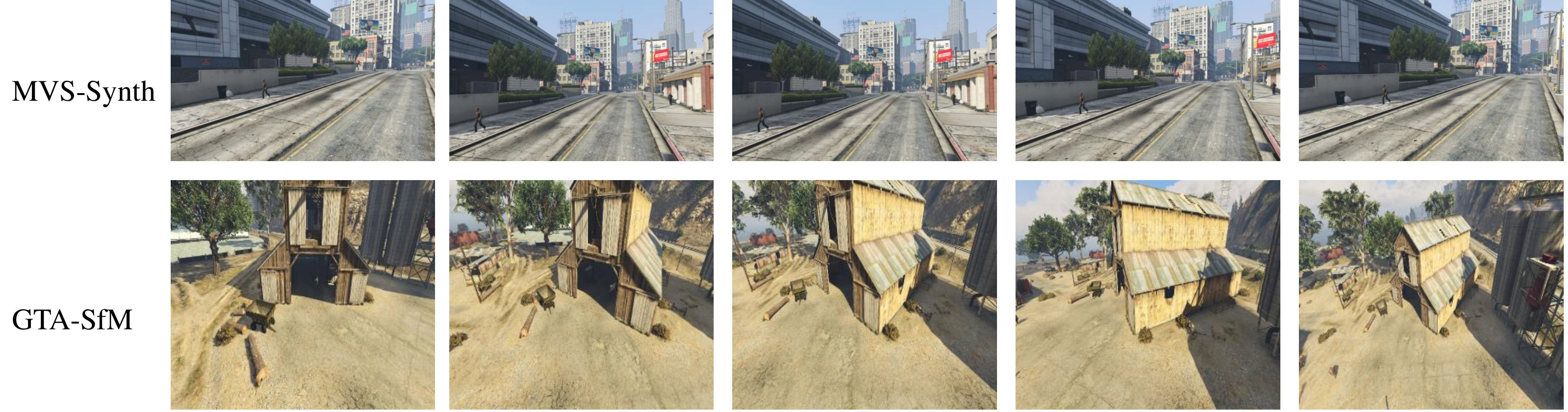}
\end{center}
\vspace{-0.5cm}
\caption{Compare the proposed GTA-SfM dataset and previous MVS-Synth dataset. The trajectory of the camera in the proposed dataset contains large translation and rotation.}
\label{fig:dataset_compare}
\end{figure*}

We provide samples of the dataset in Figure~\ref{fig:dataset_compare}. As shown in the figure, the proposed dataset is more similar to that of SfM applications.
 
\section{Experiments}
 
\subsection{Two-view Evaluation}
 
\textbf{Pose Estimation} Camera poses are estimated using optical flow in different resolutions. Here, we study the pose estimation quality together with the corresponding optical flow quality. The result is shown in Table~\ref{multi_pose}. At finer pyramid levels, the error of both optical flow and camera pose estimation decreases. However, contrary to the experience from classic SfM methods, the camera pose estimation is still better than SIFT even the optical flow resolution is $40\times32$. This can be explained by the dense pixel correspondences from the optical flow. 
 
\begin{table}[h]
\centering
\caption{Camera pose and optical flow estimation quality at different pyramid levels.}
\label{multi_pose}
\scalebox{0.95}{
\begin{tabular}{c|cccc}
Level & Resolution & Flow Err. & Rot. Err. & Trans. Err. \\ \hline
0     & 160$\times$128    & 3.168      & 1.491      & 8.517        \\
1     & 80$\times$64      & 3.269      & 1.566      & 8.559        \\
2     & 40$\times$32      & 3.707      & 1.776      & 10.630       \\ \hline
\end{tabular}}
\end{table}
 
\begin{figure*}[h]
\vspace{-2.5cm}
\begin{center}
\includegraphics[width=1.0\linewidth]{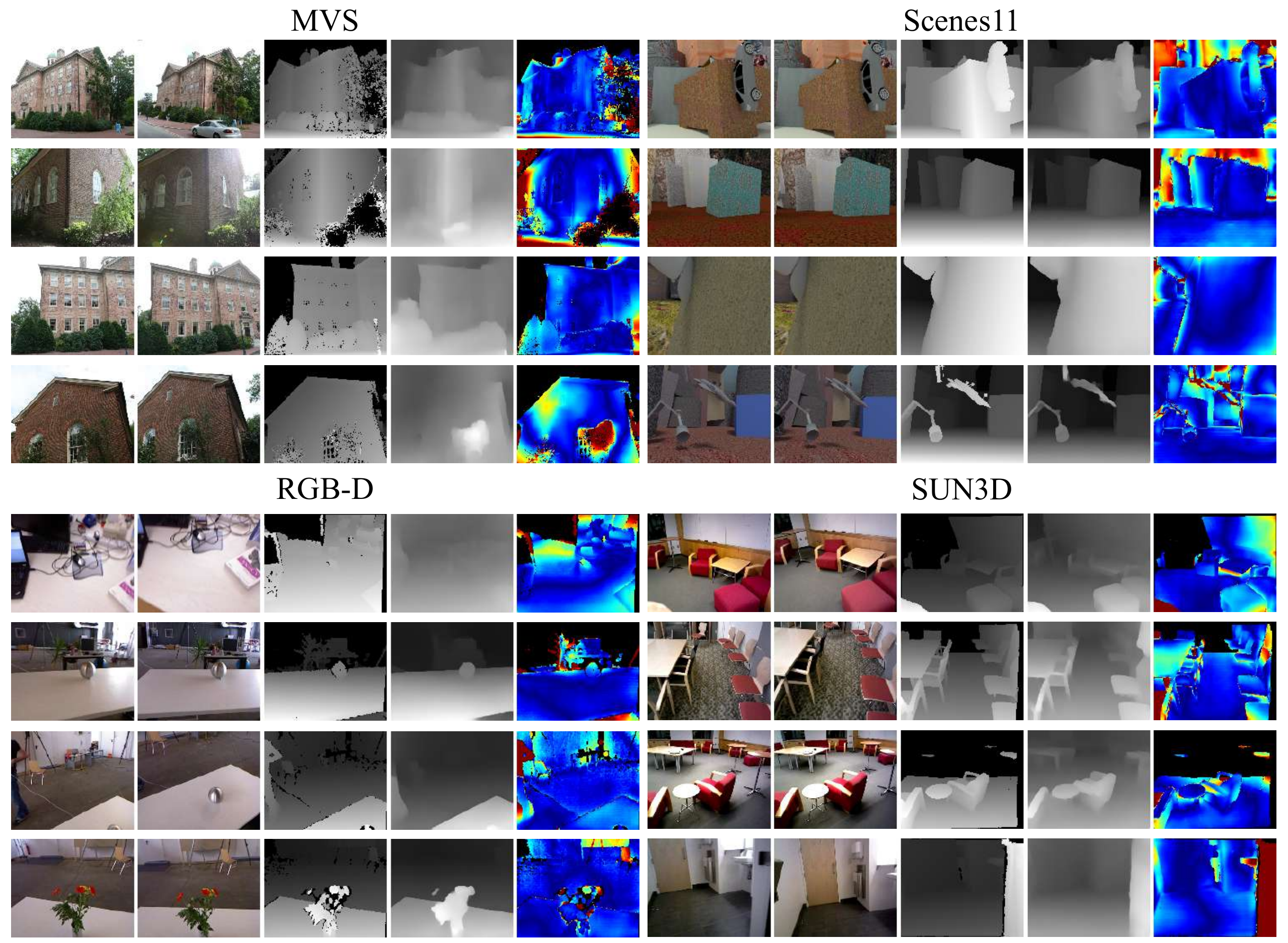}
\end{center}
\caption{Depth map estimation of the DeMoN dataset. For each sample, from left to right: source image, target image, ground truth depth map, estimated depth map, and the L1-rel error map.}
   
\label{fig:sup_demon}
\end{figure*}

\textbf{Depth Estimation} As shown in Figure~\ref{fig:sup_demon}, we provide more qualitative results of our method on the DeMoN dataset. In MVS sequence, the depth net has difficult estimating the structure of trees such as the second, and the fourth row. These trees usually have very complex structures and introduce occlusions. Such structures and are difficult even for offline methods (the ground truth also misses the trees).
 
\subsection{Depth Fusion Evaluation}
 
\begin{figure*}[h]
\begin{center}
\includegraphics[width=1.0\linewidth]{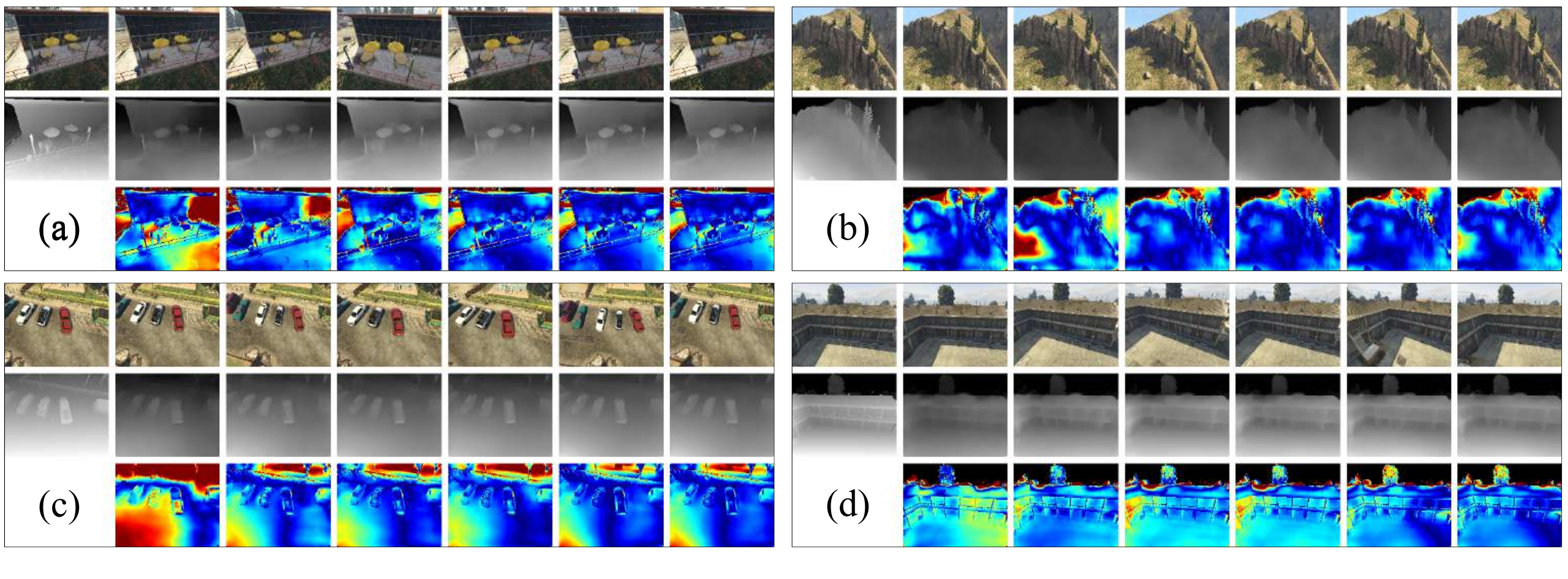}
\end{center}
\vspace{-0.5cm}
\caption{Depth map quality given different numbers of target images. For each sample, the up row shows the source image $I_s$, and five target images (from $I_{t,1}$ to $I_{t,5}$). The middle row shows the ground truth depth map and estimated depth maps with different target images. The bottom row shows the L1-rel error maps of the estimation. For example, from left to right, the fourth column shows target image $I_{t,3}$, estimated depth map which uses $I_s$, $I_{t,1}$, $I_{t,2}$, and $I_{t,3}$ as input, and the corresponding error map.}
\label{fig:gta_multiview}
\end{figure*}

\textbf{Qualitative Results} Multiview images bring more structure information of the scene, thus the depth map estimation can be robust and accurate. Figure~\ref{fig:gta_multiview} illustrates the quality of the depth estimation given different numbers of target images. The depth estimation improves when more target images are given to the depth network. Take the sample (b) and (d) as examples, the fine structures of trees and poles are well recovered by fusing multiple image pairs.
 
\textbf{Runtime Comparison} Table~\ref{multiview_runtime} shows the runtime comparison between our method and DeepMVS~\cite{deepmvs} given different numbers of target images (all measured with the same resolution). As shown in the table, our method is much more efficient compared with DeepMVS and scales well w.r.t the number of target images: from 2 target images to 6 target images, the time grows by 26\% in our method and 118\% in DeepMVS.
 
\begin{table}[h]
\centering
\caption{Estimation time comparison}
\label{multiview_runtime}
\begin{tabular}{c|ccccc}
Target Image Num. & 2  & 3  & 4  & 5 
& 6  \\ \hline
Ours (\textbf{ms})   & 49 & 53 & 53 & 57 & 62 \\
DeepMVS (\textbf{s}) & 11 & 14 & 17 & 21 & 24 \\ \hline
\end{tabular}
\end{table}
 
\begin{figure*}[h]
\begin{center}
\includegraphics[width=1.0\linewidth]{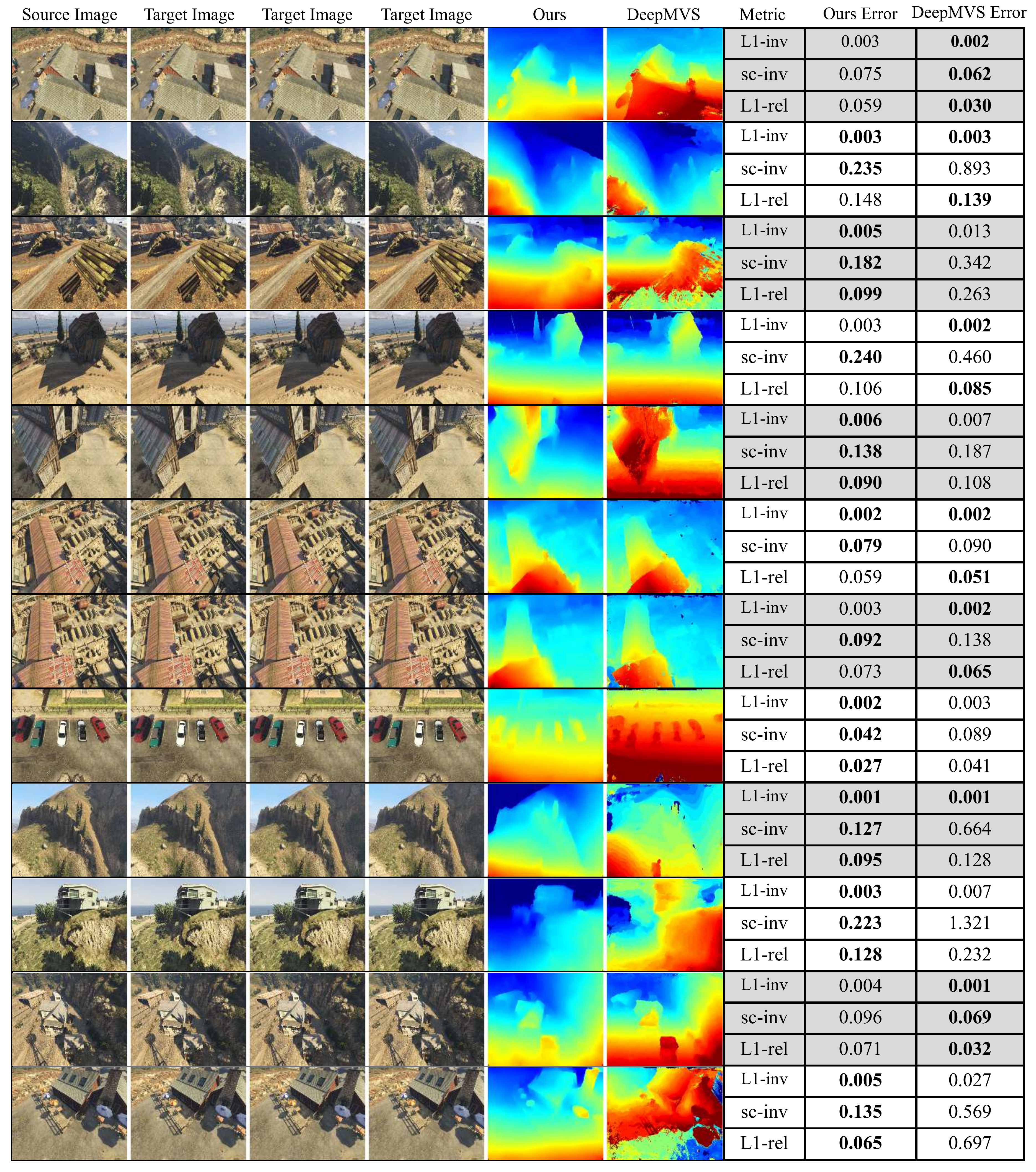}
\end{center}   
\vspace{-0.5cm}
\caption{Depth comparison between our method and DeepMVS using \textit{randomly} sampled images. The right side is the calculated depth error. Best results are highlighted in bold. As shown, our method estimates smooth and detailed depth maps, and is much more efficient than DeepMVS.}
\label{fig:compare_mvs}
\end{figure*}

\textbf{Quantitative Results} Image pairs are \textit{randomly} sampled to compare the performance of our method and DeepMVS. Each source image is observed by three target images. DeepMVS is provided with ground truth camera poses and takes more time to estimate depth maps. The results are shown in Figure~\ref{fig:compare_mvs}. At the right side of each sample, we calculate the L1-inv, sc-inv, L1-rel error of the estimated depth maps. Since depth maps from DeepMVS contain many outliers, we remove the maximum and minimum disparities before the evaluation. As shown in the figure, our method estimates smooth depth maps and is more accurate in most of the cases.
 
\begin{figure*}[h]
\begin{center}
\includegraphics[width=0.9\linewidth]{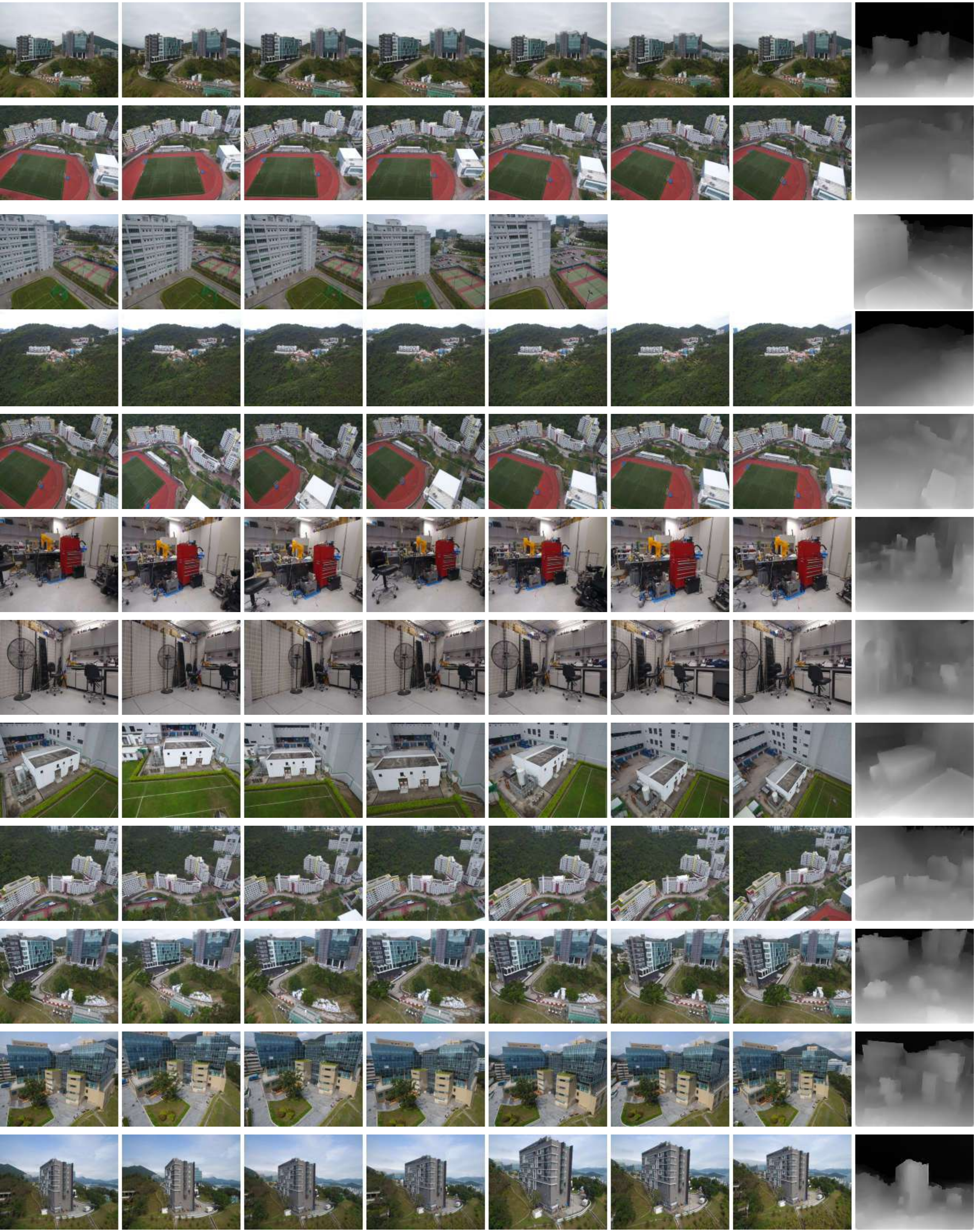}
\end{center}
\caption{Applying the GTA-SfM trained models to aerial photographs and indoor images. From left to right in each row is the source image, target images, and the estimated depth map. The source image on the third row is observed by four target images, and other source images are observed by six target images.}
\label{fig:dji}
\end{figure*}

\begin{figure*}[h]
\begin{center}
\includegraphics[width=1.0\linewidth]{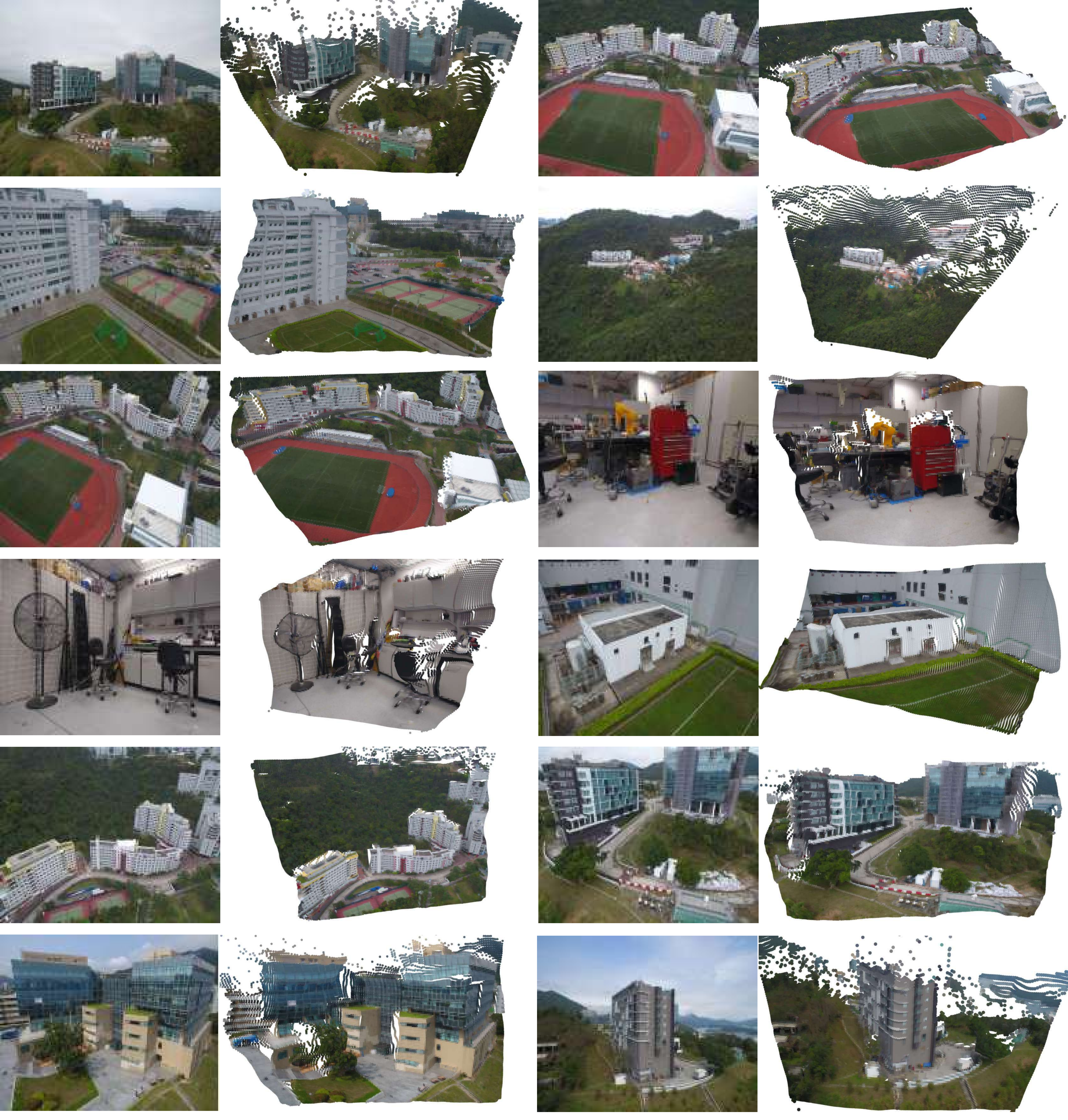}
\end{center}
\caption{Corresponding point cloud visualization of the estimated depth maps in Figure~\ref{fig:dji}. In each sample, left is the source image and right is the rendered point cloud. Pixels with depth larger than $200$ are considered as the sky and not visualized.}
\label{fig:dji2}
\end{figure*}

\begin{figure*}[h]
\begin{center}
\includegraphics[width=1.0\linewidth]{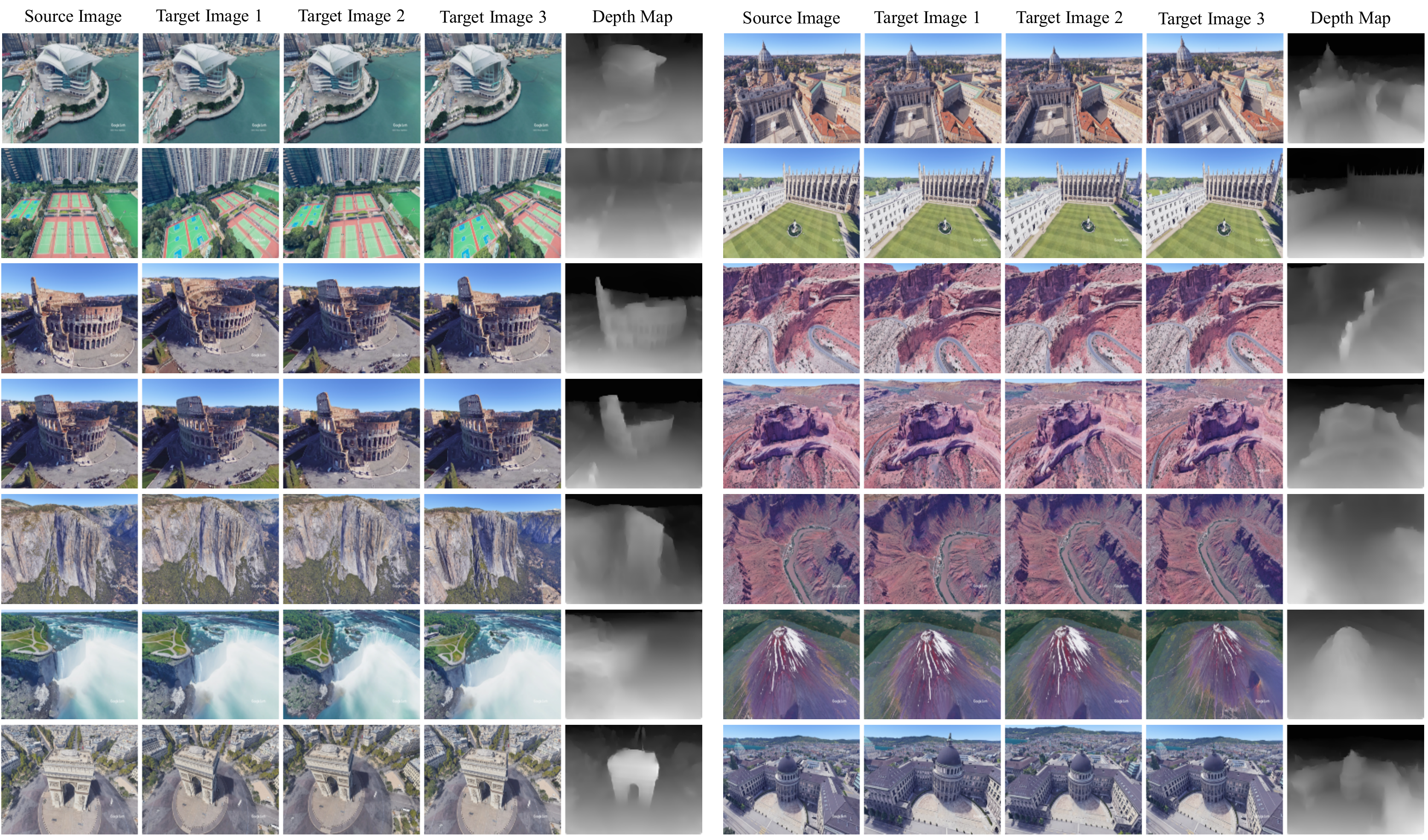}
\end{center}
\caption{Applying the GTA-SfM trained models to reconstruct images from Google Earth. Different scenes are used to show the generalization ability of the proposed method.} 
\label{fig:google_earth}
\end{figure*}

\begin{figure*}[h]
\begin{center}
\includegraphics[width=0.9\linewidth]{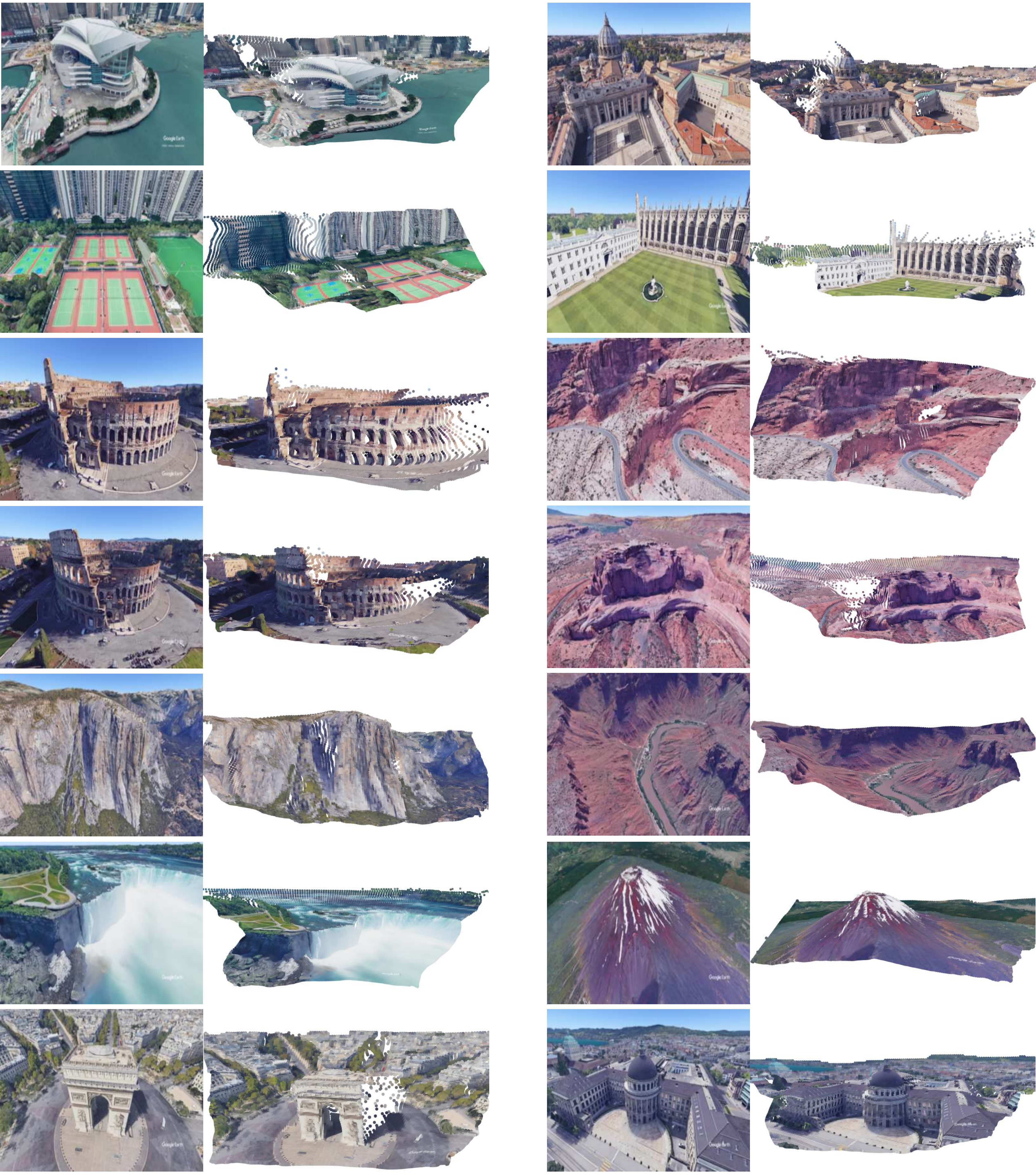}
\end{center}
    \caption{Point cloud visualization of reconstructed images from Google Earth.}
\label{fig:pointcloud}
\end{figure*}

\textbf{Generalization Ability} To demonstrate the generalization ability of the model and the effectiveness of the proposed GTA-SfM dataset, we apply the GTA-SfM trained models directly to images from real worlds and Google Earth. Figure~\ref{fig:dji} and Figure~\ref{fig:dji2} shows the estimated depth maps and point clouds of aerial photographs and indoor images. In Figure~\ref{fig:google_earth} and Figure~\ref{fig:pointcloud}, the trained model is further applied to images collected in Google Earth. Both architectures and natural scenes from different locations are covered in the experiment. As shown in the figures, even trained with synthetic images, our method can estimate depth maps from unseen scenes.
  
\section{Advantage and Limitation}
Key to the proposed method is the carefully designed flow-motion network. The high-quality optical flow and camera motion enable accurate and efficient depth triangulation. On the other hand, many prior works (e.g., LS-Net~\cite{ls_net} and CodeSLAM~\cite{code_slam}) estimate the depth maps and camera poses by iteratively minimizing the reprojection error. Such refinements are prone to local minimums and brightness changes in the images. We have demonstrated that the proposed method generates accurate camera poses and depth maps with less forward-time.
 
\begin{figure*}[h]
\centering
\includegraphics[width=1.0\linewidth]{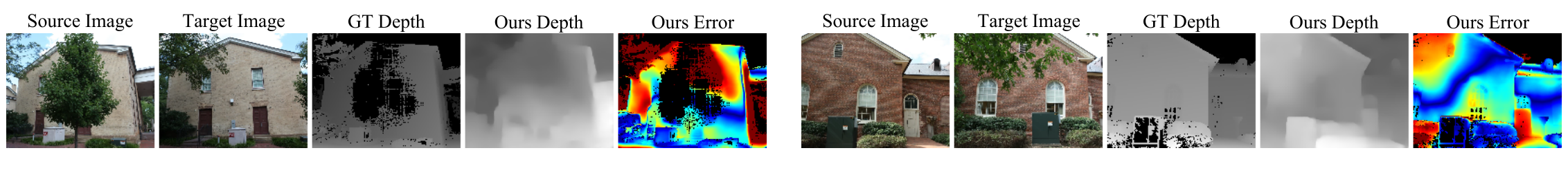}
\caption{Occlusion problem in the MVS sequence. Our method utilizes the multiview observations to triangulate the depth of each pixel thus the depth of occluded parts is difficult to be estimated. Even `ground truth' method cannot deal with such occlusion problems.}
\label{fig:occlusion}
\end{figure*}

Although achieving state-of-the-art results, the proposed method relies on high-quality optical flow estimation thus occlusion is challenging for the method. In Figure~\ref{fig:occlusion}, we show the occlusion problem in the MVS sequence. Another reason that makes such complex occlusion difficult for the network is the supervision missing from the ground truth depth maps, which also motivates us to propose the GTA-SfM dataset such that networks can be correctly trained and evaluated.
 
{\small
\bibliographystyle{unsrt}
\bibliography{egbib}
}


